# Object detection with multimodal large vision-language models: An in-depth review

Ranjan Sapkota *, Manoj Karkee *

*Cornell University, Biological & Environmental Engineering, Ithaca, 14850, NY, USA*

ABSTRACT

The fusion of language and vision in large vision-language models (LVLMs) has revolutionized deep learning-based object detection by enhancing adaptability, contextual reasoning, and generalization beyond traditional architectures. This in-depth review presents a structured exploration of the state-of-the-art in LVLMs, systematically organized through a three-step research review process. First, we discuss the functioning of vision language models (VLMs) for object detection, describing how these models harness natural language processing (NLP) and computer vision (CV) techniques to revolutionize object detection and localization. We then explain the architectural innovations, training paradigms, and output flexibility of recent LVLMs for object detection, highlighting how they achieve advanced contextual understanding for object detection. The review thoroughly examines the approaches used in integration of visual and textual information, demonstrating the progress made in object detection using VLMs that facilitate more sophisticated object detection and localization strategies. Furthermore, this review presents comprehensive visualizations demonstrating LVLMs' effectiveness in diverse scenarios including localization and segmentation, and then compares their real-time performance, adaptability, and complexity to traditional deep learning systems. Based on the review analysis, its is expected that LVLMs will soon meet or surpass the performance of conventional methods in object detection. However, because of the unique and complimentary characteristics of traditional deep learning approaches and LVLMS, it is anticipated that hybrid approaches integrating both types of object detection models will be utilized in the future to maximize the speed, reliability and robotiness of the systems. Moreover, the review also identifies a few major limitations of the current LVLM modes, proposes solutions to address those challenges, and presents a clear roadmap for the future advancement in this field. We conclude, based on this study, that the recent advancement in LVLMs have made and will continue to make a transformative impact on object detection and automated applications in the future.

## 1. Introduction

### 1.1. Background

Object detection is a crucial component of machine vision systems that identifies and locates objects within images or videos, enabling machines to intelligently interact with their surroundings [17]. Efficient and accurate object detection plays a crucial role in monitoring and automating various tasks/operations in a wide range of industries. For instance, in autonomous vehicles, accurate object detection and localization facilitate safe navigation by detecting pedestrians, vehicles, and road signs [18] where as in healthcare, detecting anomalies like tumors in medical scans plays a critical role for timely and accurate diagnostics [19]. In retail, supporting automated inventory management [20] is possible with accurate objective identification whereas in agriculture, accurate object detection helps enhance precision farming by monitoring crop health and detecting pests [21]. Security and surveillance is another important industry relying on improved detection of unauthorized activities such as access to homes and businesses [22].

Historically, as illustrated in Fig. 1, prior to the advent of deep learning (DL), object detection relied on methods like Background Subtraction [1,23], which differentiates moving objects from static backgrounds but struggles with dynamic scenes. Similarly, Haar Cascades [2,24] was another approach that detect faces through cascade stages but is not robust against orientation and scale variations. Similarly, Histogram of Oriented Gradients (HOG) [3,25] technique was quite widely used but is sensitive to orientation and lighting; whereas

* Corresponding authors.
*E-mail addresses:* rs2672@cornell.edu (R. Sapkota), mk2684@cornell.edu (M. Karkee).

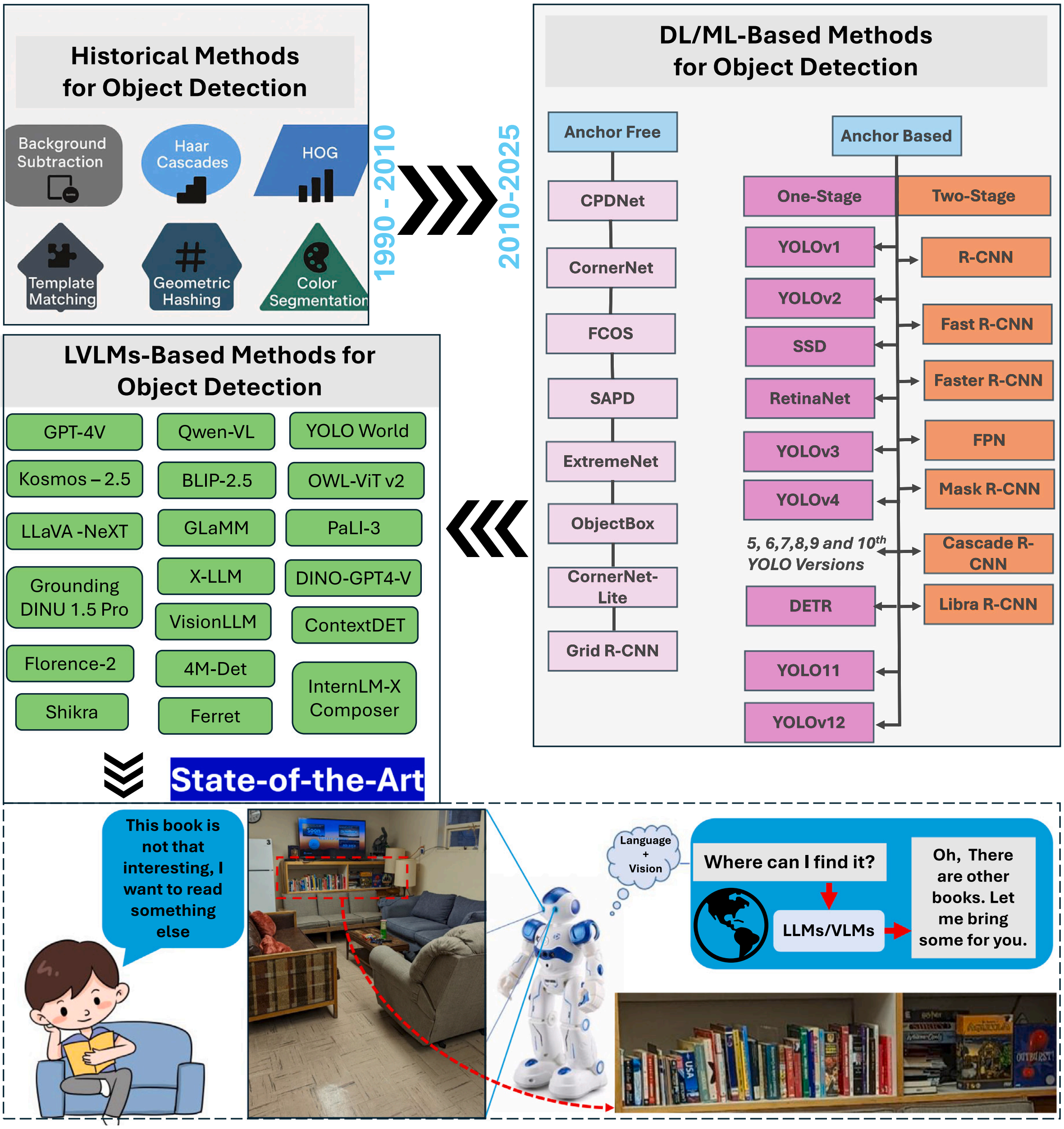

**Fig. 1.** Comprehensive illustration of the evolution of object detection methodologies from conventional techniques to advanced Large Vision Language Models (LVLMs). Historically, object detection methods such as Background Subtraction [1], Haar Cascades [2], Histogram of Oriented Gradients (HOG) [3], and Template Matching [4] laid foundational principles. Transitioning to deep learning (DL) and machine learning (ML), significant advances were made through methods such as SSD [5], YOLO (You Only Look Once) [6], Faster R-CNN [7], Mask R-CNN [8], RetinaNet [9], and EfficientDet [10], which revolutionized speed and accuracy in detection tasks. More recently, LVLMs such as ContextDET [11], VOLTRON [12], DVDet [13], DOD Framework [14], Synthetic negative generation [15], and DetGPT [16] have integrated complex language understanding capabilities, enabling dynamic and contextually aware object detection across diverse and challenging environments. Object detection has advanced from simple methods to complex vision-language models, enabling machines to understand and interact with their surroundings more effectively. These models interpret contextual cues for more accurate, practically applicable real-world detections, merging language and vision to increase detection capabilities. The figure illustrates a futuristic application of this concept: a person is seen reading a book but appears disinterested, expressing the thought, “This book is not that interesting, I want to read something else”. A robot equipped with a vision-language model perceives the situation, detects the bookshelves in its environment, understands the user’s sentiment through language processing, and identifies suitable alternative books from the shelf. This exemplifies how vision language models (VLMs) enable robots to comprehend nuanced human intent, detect relevant objects, and respond accordingly, showcasing the emergence of general intelligence in object detection through multimodal perception and reasoning.

Template Matching [4,26] has been limited by scale and rotation changes. Another approach used in the past is Geometric Hashing [27], which is memory-intensive and sensitive to noise whereas Color Segmentation [28,29] has been affected by lighting variability and similar colors between objects and backgrounds.

To summarize, these methods of object detection, predominantly developed or applied from 1990 to 2015, achieved limited success but laid the foundation for today’s advanced techniques. These historical methods such as HOG, often struggled with variations in object orientation, scale, and lighting conditions, limiting their effectiveness in dynamic or complex environments [30,31]. Additionally, techniques like Background Subtraction and Color Segmentation were particularly susceptible to changes in background dynamics and lighting, making them unreliable for consistent object identification across varying scenarios [32,33].

Over the past 15 years, ML/DL [34] have quickly transformed object detection (Fig. 1) tasks, introducing a number of sophisticated models that significantly surpass the capabilities of traditional methods [35]. For instance, Single Shot MultiBox Detector (SSD) [5] efficiently processes images in one shot to detect objects, delivering both their locations and class predictions. Likewise, YOLO streamlines detection by dividing images into grids, each predicting bounding boxes and probabilities, enabling rapid real-time detection [6,36]. Additionally, Fast R-CNN [7] and Faster R-CNN [7] enhance detection by using region proposal networks and shared convolutional features, respectively, to quickly and accurately predict object locations and classes [37].

Additionally, Mask R-CNN [8] builds on Faster R-CNN by adding a segmentation overlay that provides precise pixel-level object outlines, while RetinaNet uses a focal loss to focus on hard-to-detect objects, balancing the detection of various object sizes [9,38]. Furthermore, EfficientDet combines efficient scaling and bi-directional feature networks, optimizing speed and scalability in object detection without sacrificing accuracy [10].

Recent models like RT-DETR [39] and RTMDet [40] further advance real-time detection, with notably outperforming a few traditional YOLO metrics in some scenarios. Grounding DINO represents a cutting-edge development in zero-shot detection [41], capable of identifying objects without prior specific training on their classes [42]. Likewise, other innovative approaches include SqueezeDet [43], tailored for autonomous driving, and MobileNet, designed for mobile applications due to its lightweight architecture [44,45]. In addition, CenterNet marks a shift from traditional bounding box methods by detecting objects at their central point, simplifying the detection mechanism [46]. Cascade R-CNN, on the other hand, iteratively refines detections, enhancing accuracy through multiple stages [47,48]. In the domain of transformers, Vision Transformer (ViT) [49,50], and Swin Transformer [51,52] have been adapted for object detection, leveraging the transformer architecture to enhance contextual understanding significantly. PP-YOLOE [53], YOLO11 and YOLOv12 [54] are recent iterations in the YOLO family, improving generalization and performance across diverse detection tasks.

Although DL methods for object detection, such as YOLO, R-CNN, and SSD, have made significant progress in machine vision, they face several limitations and challenges. These models often require extensive labeled datasets for training, which can be time-consuming and expensive to create [55]. They may struggle with zero-shot learning [56], making it difficult to detect objects not present in the training data [57]. DL models can also be computationally intensive [58], especially for real-time applications [59]. Their performance can degrade when dealing with small objects, occluded objects, or complex scenes with multiple overlapping items [17]. Additionally, these models may lack the contextual understanding necessary for object interpretation in varied environments [60]. They typically provide bounding boxes and class labels but struggle with more detailed descriptions or answering queries about the detected objects [61]. Furthermore, fine-tuning these models for specific domains or new object classes often requires significant expertise and computational resources, limiting their adaptability in rapidly changing or specialized applications [62].

Following the limitations of traditional deep learning models in object detection, the emergence of LVLMs marks an important shift in the field, positioning them as state-of-the-art methodologies (as depicted in Fig. 1). Unlike conventional models that operate solely on visual input, multimodal LVLMs process and integrate various data modalities such as text, images, and even video, enabling a more comprehensive and semantically rich understanding of visual scenes [63,64]. LVLMs are designed to bridge visual recognition with natural language understanding, allowing them to interpret images, generate relevant descriptions, and answer contextually grounded questions. This multimodal capacity enables these models to detect and classify objects not just by appearance but also by their contextual relationships [65,66].

A key advantage of LVLMs is their ability to generalize to unseen classes through zero-shot learning, identifying objects that were not explicitly present in their training datasets [67–69]. Furthermore, while LVLMs are generally computationally intensive, recent adaptations have introduced more efficient variants that strike a balance between accuracy and latency, enabling deployment in real-time applications that demand immediate perception and action [11,64]. The bottom panel of Fig. 1 illustrates a futuristic real-world application of LVLM-based object detection. In this scenario, a person is shown reading a book and expresses disinterest by thinking or prompting, “This book is not that interesting, I want to read something else”. A robot equipped with vision-language capabilities interprets this language input, scans the surrounding bookshelf, detects the relevant books using object detection, and responds with an appropriate suggestion. This scene exemplifies the current state-of-the-art in object detection: real-time, and context-aware interaction driven by the integration of vision and language, highlighting the general intelligence potential of multimodal LVLM systems.

In this review, we present the first comprehensive examination of object detection methodologies using multimodal LVLMs, covering the advancements from 2022 to 2025. We investigate the architectural and operational features of leading systems such as GPT-4V, LLaVA-1.5, and SpatialLM, and compare their performances with those achieved with traditional deep learning models like YOLO, SSD, and Faster R-CNN. While conventional methods prioritize bounding box accuracy and inference speed, LVLMs offer enhanced semantic reasoning and adaptability through cross-modal learning, enabling zero-shot detection and improved contextual understanding in complex environments. We also analyze the limitations of LVLMs, including their challenges in precise spatial localization, and emphasize the need for hybrid frameworks that fuse the contextual intelligence of LVLMs with the spatial precision of conventional object detectors.

Beyond the performance analysis, this review explores how LVLMs are transforming vision tasks through natural language interfaces, addressing critical issues of computational efficiency, deployment feasibility, and domain-specific adaptability. We assess their industrial applications, compare their trade-offs with established models, and propose future research directions. In summary, this review serves as a foundational study that synthesizes key capabilities, challenges, and practical strategies for implementing LVLMs in object detection, establishing a baseline for ongoing and future advancements in this rapidly evolving field.

### 1.2. *Review methodology*

#### 1.2.1. *Review motivation and structure*

This review aims to provide a comprehensive synthesis of object detection using multimodal LVLMs, focusing on their architectures, training foundations, performance characteristics, and practical applicability across diverse detection settings. The scope encompasses recent studies from 2022 to 2025, with an emphasis on models that integrate vision-language fusion for object understanding. The structure of the review is guided by three core research questions (RQs), shown in Fig. 3, which inform our comparative and analytical approach across traditional and multimodal systems.

The paper selection and filtration process used in this review is summarized in Fig. 2a. This systematic approach ensured that the most relevant and technically grounded contributions were included across subdomains. To illustrate the accelerating momentum in this research area, Fig. 2b visualizes the year-wise distribution of reviewed works. Notably, the number of qualifying studies rose from just six in 2022 to fifty in the early months of 2025, underscoring the rapidly growing importance of LVLMs in object detection.

#### 1.2.2. *Literature discovery and filtering strategy*

To ensure comprehensive and methodologically sound coverage, we conducted a systematic literature search across a diverse set of reputable academic databases and AI-centric platforms, including both peer-reviewed repositories and preprint servers (e.g., IEEE Xplore, Web of Science, arXiv) as well as community-driven hubs like Hugging Face and ChatGPT. The initial search phase used broad keywords such as *"object detection"*, *"vision-language models"*, and *"large language models"* to capture the evolving landscape of multimodal detection systems.

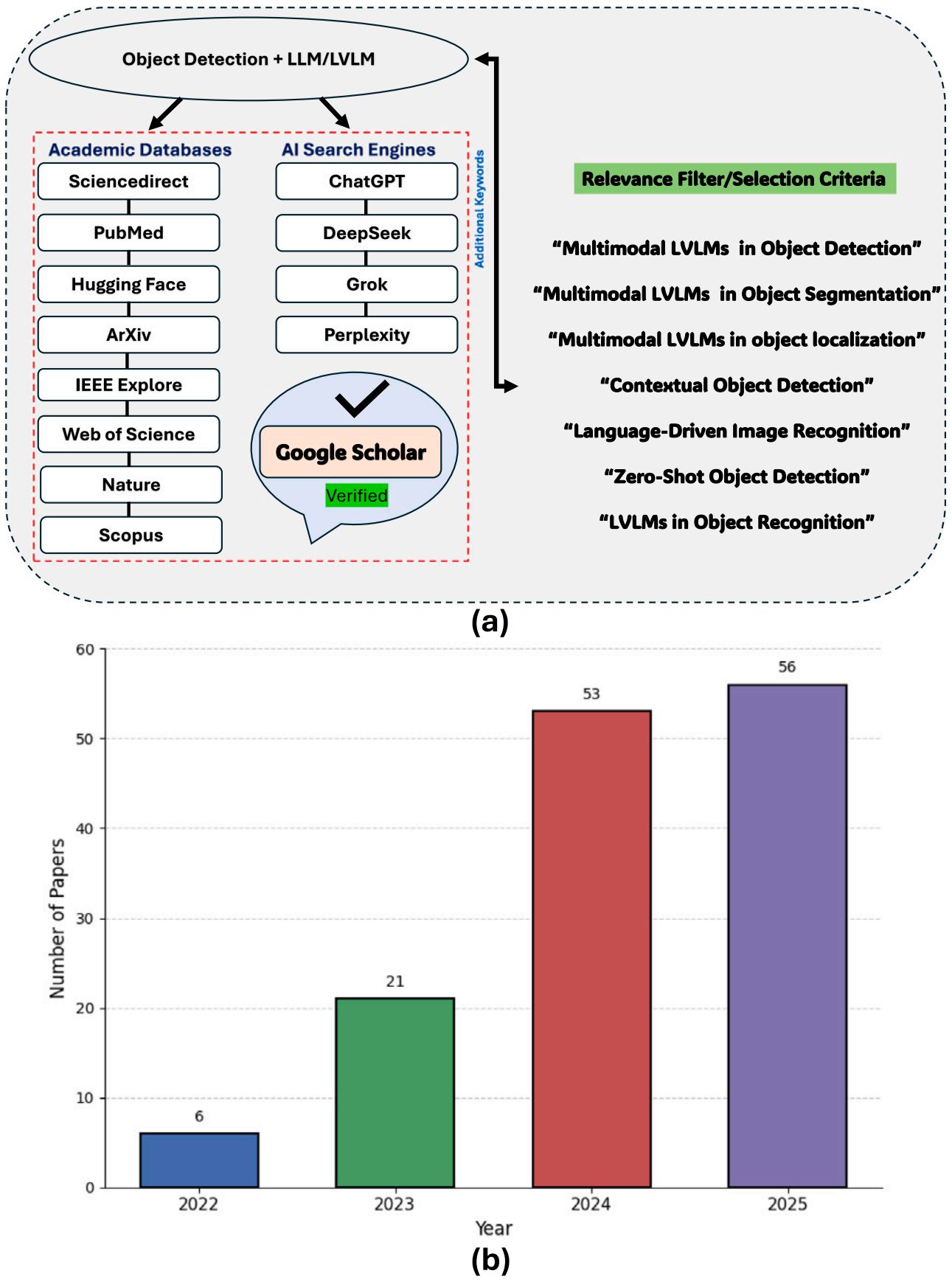

**Fig. 2.** (a) Streamlined search and filtering process applied for paper selection in this review, using twelve search engines and refined multimodal keyword combinations; (b) Temporal distribution of reviewed papers shows rapid growth in publications using LVLMs for object detection (as of April 20, 2025).

Subsequent refinement employed task-specific and domain-sensitive terms such as *"multimodal LVLMs"*, *"prompted object localization"*, and *"image-text grounding"* to isolate relevant contributions. Inclusion criteria emphasized models that incorporated pretrained or fine-tuned vision-language architectures applied to object detection tasks. Studies were evaluated for architectural transparency, methodological rigor, and relevance to either foundational development or real-world deployment. Works lacking sufficient technical depth, or relying solely on black-box APIs without reproducible methodologies, were excluded. The final corpus reflects a curated synthesis of impactful research spanning model design, evaluation strategies, and deployment contexts in LVLM-based object detection (Fig. 2).

#### *1.2.3. Review design and research questions*

This review is organized around three central RQs that define the thematic and technical boundaries of our analysis. A conceptual overview of these RQs is illustrated in Fig. 3.

1. **Foundational Functioning & Evolution of Multimodal LVLMs in Object Detection:** What novel capabilities and representational mechanisms do LVLMs bring to object detection? How do they handle multimodal fusion, segmentation, and scene understanding?

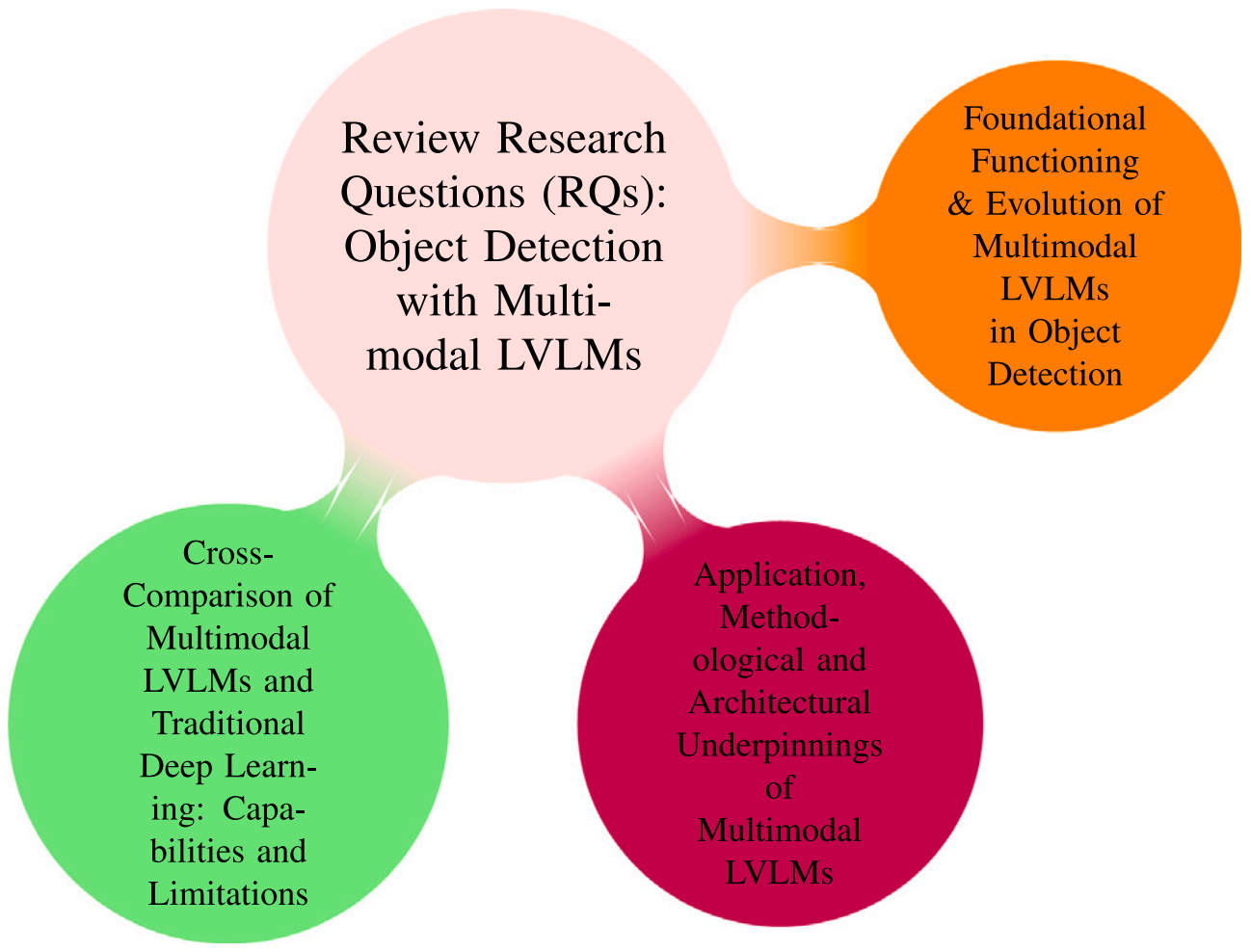

**Fig. 3.** Conceptual structure of this review organized around three research questions guiding analysis of LVLMs in object detection.

2. **Methodological and Architectural Underpinnings:** What are the architectural choices (e.g., encoders, tokenizers, alignment modules) and training strategies that define state-of-the-art LVLMs in detection tasks?
3. **Cross-Comparison with Traditional Deep Learning:** How do LVLMs compare with classic models like YOLO, SSD, and Faster R-CNN in terms of detection performance, generalization, efficiency, and deployment readiness?

Throughout this review, we systematically analyze LVLM performance across diverse datasets, object granularities, environmental conditions, and inference constraints. In response to reviewer feedback, we have incorporated detailed quantitative performance comparisons evaluating mAP, zero-shot accuracy, and inference speed across both emerging LVLMs and traditional deep learning baselines. These assessments, along with architectural analysis, provide a nuanced understanding of each model's real-world usability and trade-offs.

The remainder of this paper is structured to address the three core research questions (RQs), with each section examining the current challenges, limitations, and emerging solutions in LVLM-based object detection. We analyze model performance across benchmark datasets, object scales, environmental variability, and inference conditions. Particular emphasis is placed on architectural strengths and real-world usability, enabling a rigorous comparison with conventional deep learning models. Additionally, we explore the implications of real-time LVLM-based detection in robotic systems, highlighting how these models advance perception, decision-making, and adaptability in dynamic environments across agricultural, industrial, and general-purpose automation domains.

## 2. Foundational functioning and evolution of multimodal LVLMs in object detection:

Historically, LVLMs were generally pre-trained from scratch, thus building models entirely from raw data without leveraging pre-existing language or vision models [70,71]. This approach required simultaneous training of both visual and linguistic components on massive multimodal datasets [72]. For instance, early LVLMs like Flamingo were trained from scratch using extensive resources such as 2.3 billion web pages and 400 million image-text pairs [73]. This method involved starting models with random weights, which required them to learn language understanding, visual processing, and cross-modal alignment all at once. The key aspects of training a LVLM from scratch include:

**Table 1**
Comparative analysis of LVLM training approaches highlighting differences in efficiency, text task performance, data requirements, and architectural flexibility between scratch-trained models and those leveraging pre-trained LLM backbones.

| Aspect | From-scratch LVLMs | Pre-trained LLM-based LVLMs |
|---|---|---|
| Training efficiency | 20%–50% slower convergence[a] | Faster adaptation via frozen LLM layers [82] |
| Text task performance | 15% drop on language benchmarks [82] | Preserves LLM's original capabilities [82] |
| Data requirements | 10–100x more multimodal data [83] | Works with smaller domain datasets[b] |
| Architectural flexibility | Rigid end-to-end design [83] | Modular visual adapter layers [82] |

[a] https://fritz.ai/pre-trained-machine-learning-models-vs-models-trained-from-scratch/.
[b] https://magazine.sebastianraschka.com/p/instruction-pretraining-llms.

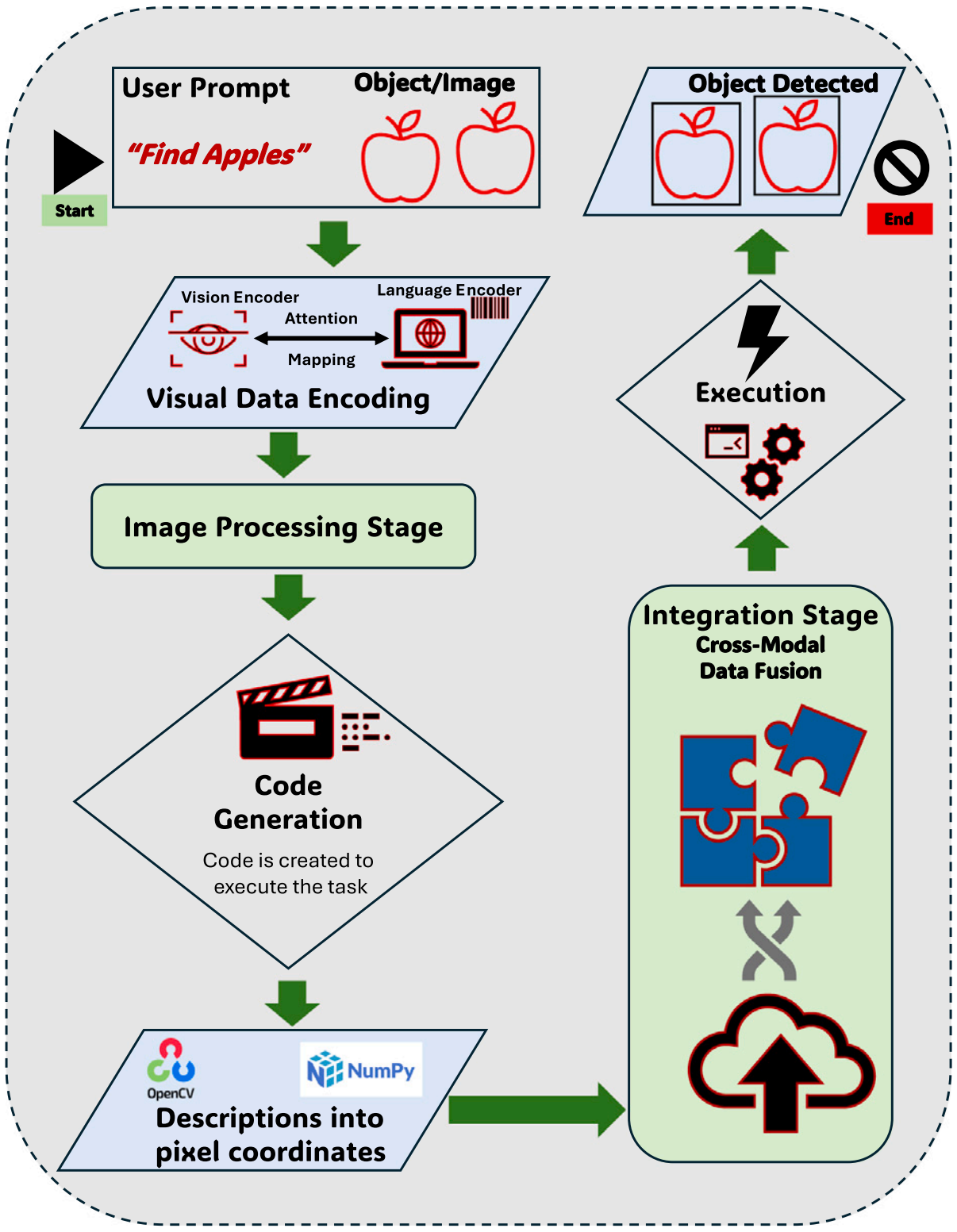

**Fig. 4.** Illustration of the object detection process with multimodal LVLMs, which starts with raw image data that is transformed into structured visual embeddings. Textual prompts align with these images, and cross-modal fusion enhances contextual understanding for accurate object localization, resulting in comprehensive detection outputs.

- **Full Initialization:** Models begin with random weights, which does not inherit any knowledge from existing LLMs or vision models [74].
- **Data Requirements:** This process depends on extremely large multimodal datasets for developing a reliable and accurate model [11,75].
- **Challenges:** Building LVLM models from scratch requires high computational costs (often involving months of GPU training) [76,77] and poses risks like "catastrophic" forgetting where a model may lose previously learned information as it acquires new, potentially because of conflicting data [78,79]. Moreover, these models often experienced performance degradation on text-only tasks when compared to those using LLM backbones [80].

Recent trends, however, show a strategic shift (Table 1) towards utilizing pre-trained LLMs as foundational backbones, followed by adding visual modules through efficient fine-tuning [81–83]. This adaptation (as illustrated in Fig. 4) facilitates a more seamless alignment between visual inputs and textual data, illustrating the LVLM's ability to interpret and process multimodal information more comprehensively.

Fig. 4 demonstrates the technical workflow in LVLMs using apple detection as an example to illustrate the stages of object detection and localization. Vision encoders such as CLIP [84] and BLIP [85], trained on vast multimodal datasets, are key in capturing intricate visual-textual relationships, enhancing multimodal understanding when paired with LLMs. This integration, including the use of advanced text encoders such as in ALIGN [73] and LLaVA [86], marks a significant shift in Vision-Language Model (VLM) architectures, enhancing their efficiency and adaptability for complex tasks by treating visual features as tokens and enabling seamless, dynamic cross-modal interactions.

This process of object detection and localization with multimodal LVLMs can be summarized in seven key steps as follows:

1. **User Prompt:** The process begins with a user input, such as "find apples". This simple, human-like interaction initiates the LVLM's processing sequence, combining natural language understanding with visual data analysis. The interaction acts as a bridge, merging linguistic queries with visual search tasks.
2. **Visual and Language Encoding:** A visual data encoder analyzes the image to extract relevant features, which are then synchronized with the textual prompt through a language encoder [87,88]. An attention mechanism facilitates this alignment by correlating the text "find apples" with corresponding regions in the image, ensuring focus on relevant visual cues that match the textual description [73,89].
3. **Code Generation:** In some recent studies, LVLMs have been utilized not only for direct perception tasks to detect objects but also as agents capable of reasoning through multimodal prompts to generate task-specific executable code or pseudo-code [90,91]. Importantly, it is noted that this code generation does not imply generating object detection algorithms from scratch. Instead, the LVLM interprets natural language queries and visual context to dynamically synthesize or select small code snippets (e.g., for drawing bounding boxes, querying object attributes, or controlling downstream modules). This capability is especially useful in tool-augmented or embodied AI settings, where LVLMs interact with external tools or APIs (e.g., for visualization or robotic control) [92,93]. In these cases, code generation serves as an intermediate reasoning step, enhancing interpretability and modular task execution. Thus, in this fashion, LVLMs operate not only as a perception model but also as a cognitive planner, translating user intent and scene context into structured actions via code generation that bridges the gap between language, vision, and programmable outputs.
4. **Conversion to Actionable Data:** Tools such as OpenCV, NumPy, and other machine learning libraries transform attention maps into precise pixel coordinates for multimodal LVLMs-based object detection [90,94]. This crucial transition converts high-level, model-generated insights into concrete, actionable data points, which are essential for creating accurate bounding boxes around the detected objects.
5. **Integration and Execution:** The visual and textual data are further synthesized during the integration stage to refine object localization [93]. In the execution phase, the model output is interpreted often as tokenized code or structured instructions and is processed to accurately identify and localize the target object within the image.

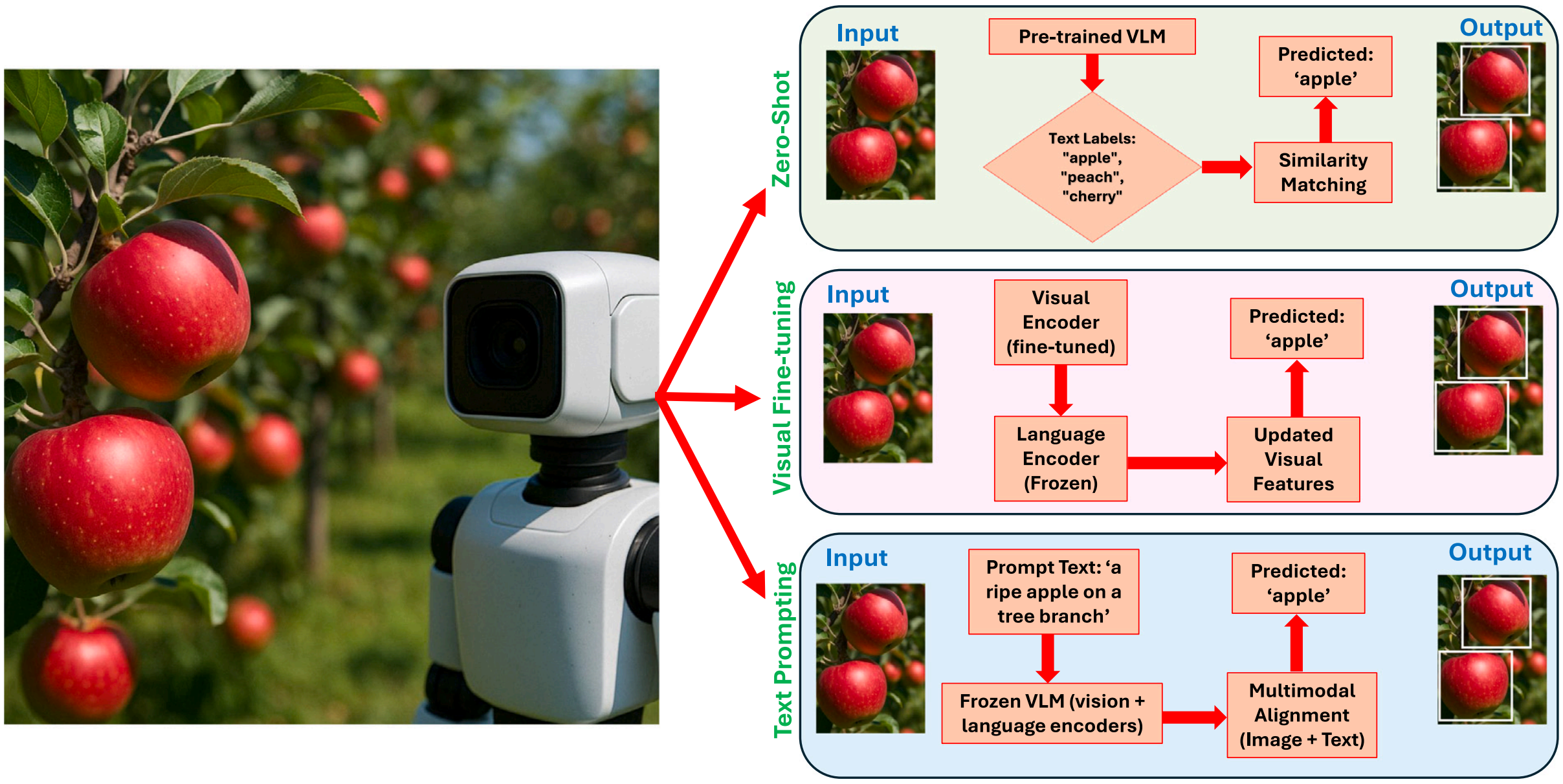

**Fig. 5.** An illustration of three LVLM-based object detection strategies: Zero-Shot Prediction, Visual Fine-Tuning, and Text Prompting, demonstrating how each method processes image-text alignment for accurate object recognition.

6. **Final Detection:** The culmination of this process is observed in the final stage of object detection, where the LVLMs not only detect but also contextually understand and present the objects within precise bounding boxes, as depicted in Fig. 4.

Application approaches or strategies for LVLM-based object detection can be categorized into three groups as illustrated in Fig. 5, each offering distinct capabilities in terms of generalization, adaptability, and supervision. To illustrate these approaches, the process of detecting an object such as an apple is used as a representative example. The three fundamental strategies are: Zero-Shot Prediction, Visual Fine-Tuning, and Text Prompting:

1. **Zero-Shot Prediction:** In this method, a pre-trained LVLM is used without any task-specific fine-tuning [95]. When the vision system receives an input image containing an object (e.g., an apple), the model evaluates the similarity between the visual features and a set of candidate textual labels such as "apple", "orange", or "cherry". Based on semantic alignment learned during pretraining, the LVLM selects the most relevant label [96,97]. This process eliminates the need for labeled training data and enables general-purpose object detection, although accuracy may vary in complex or unfamiliar domains.
2. **Visual Fine-Tuning:** In this approach, the LVLM's visual encoder is fine-tuned using a labeled dataset of domain-specific images, while the language encoder remains unchanged [98–100]. For example, if the object of interest is an apple, the visual encoder adapts to features such as shape, size, color, and occlusion commonly observed in the desired environment. This targeted fine-tuning improves object detection performance by aligning visual representations more closely with the specific context in which the image processes [101,102].
3. **Text Prompting:** This method modifies only the textual input to the LVLM, keeping both the vision and language encoders frozen [93,103]. Instead of using a basic label like "apple", descriptive prompts such as "a ripe red apple on a tree branch" are used to enhance the alignment between text and image features. These prompts guide the model to attend to the most relevant visual information without requiring any model retraining. Text prompting is a lightweight, flexible strategy especially suited for quick deployment or tasks with limited labeled data [95,104].

### 2.1. Advancements in multimodal LVLMs for object detection

The advancements in multimodal LVLMs for object detection, compared to traditional deep learning approaches, can be categorized into three major domains: (1) Architectural innovations, where modern LVLMs incorporate dual encoders or unified transformers to jointly process visual and linguistic information, enabling richer and more semantically relevant feature representations; (2) Training paradigms, which leverage large-scale image-text datasets and alignment objectives to facilitate efficient, context-aware learning across modalities, often improving zero-shot and few-shot detection performance; and (3) Output flexibility, which allows these models to not only produce accurate bounding boxes but also generate text-aligned object descriptions, supporting more interpretable and instruction-following detection capabilities.

These advances are driven by a growing series of models, each contributing novel mechanisms for visual relevance, language alignment, and multimodal reasoning. Fig. 6 presents a comprehensive timeline of major object detection LVLMs introduced from 2022 to the present, illustrating the field's rapid evolution and the increasing advancement on model architectures and their capabilities. This progression highlights a clear trend toward unified multimodal representations that support diverse downstream tasks beyond detection alone, including segmentation, captioning, and visual question answering.

- **Architectural Innovations in LVLMs:** Traditional deep learning models such as YOLO, SSD, and Faster R-CNN are fundamentally built on convolutional neural networks, each optimized for specific aspects of object detection [105]. YOLO, known for its single-stage detection mechanism, divides the image into grids, predicting bounding boxes and class probabilities directly from these grid cells using anchor boxes [6,106]. SSD extends this by utilizing multiple feature maps to detect objects across various scales in a single forward pass, optimizing for speed [5]. Faster R-CNN introduces a two-stage approach, initially generating region proposals through a region proposal network, then refining these proposals to precise bounding boxes and classifications [7]. These architectures excel in closed-set detection scenarios, where the object classes are predefined (e.g., 80 COCO categories), but they struggle to adapt beyond their trained categories. Multimodal LVLMs bring a transformative approach to object detection by incorporating language models that facilitate a robust integration of visual and textual data, which allows these models to interpret images not just as arrays of pixels, but as entities embedded with contextual information that can be described in natural language. Key architectural elements in such systems often include:

- **Dual-Stream Architectures:** LVLMs often feature dual-stream architectures, processing visual and textual data through separate pathways before integration. This architectural choice allows for dynamic adjustment of weights between visual and textual features, critical for tasks requiring detailed understanding [107,108].
- **Transformer-Based Design:** At the core of many LVLMs is the transformer architecture, adapted from NLP to handle mixed data types. This adaptation enables LVLMs to process images as sequences of patches and descriptions as sequences of tokens, enhancing their capability to generate contextually rich interpretations [107,108].
- **Attention Mechanisms:** LVLMs incorporate attention mechanisms that focus on relevant image parts in relation to textual descriptions. This feature is crucial for performing zero-shot object detection, where the model predicts objects that have not been seen during training [107,108].
- **Contextual Embedding Layers:** These models utilize advanced embedding techniques to create a shared high-dimensional space for visual and textual inputs. This integration enhances the mutual understanding between the modalities, leading to more accurate object detection [107,108].

- **Training Paradigm and Output Flexibility** The training paradigms of object detection with multimodal LVLMs differ significantly. Traditional models like YOLO and Faster R-CNN rely on meticulously labeled datasets, such as COCO, with bounding box annotations for specific classes [109]. Training involves thousands of images, and expanding to new classes requires collecting, annotating, and retraining with additional data, a time-consuming and resource-intensive process [110].
Conversely, multimodal LVLMs like GPT-4V are pretrained on vast, web-scale datasets of image-text pairs, learning rich visual-linguistic representations [84]. This pretraining enables zero-shot detection, where objects can be identified from natural language descriptions without class-specific annotations, offering scalability and adaptability. Traditional models such as Mask R-CNN produce detailed outputs, including pixel-level segmentation masks, but are constrained to a fixed set of predefined class labels (e.g., "car", "dog"). In contrast, multimodal LVLMs like Ferret (Ferret LLM, Los Angeles, California, USA) surpass these limitations by generating rich, free-form textual descriptions (e.g., "the red car near the tree") and even providing spatial outputs, such as bounding box coordinates, directly through language responses. This expanded expressiveness enables more dynamic understanding and interaction with visual scenes beyond rigid label constraints [111,112].
Additional details on the training data and paradigm of recent multimodal LLMs is presented in Table 2.

- **Contextual Understanding:** A critical advantage of multimodal LVLMs like Qwen-VL over traditional models such as YOLO and SSD is their superior contextual understanding. Traditional models are adept at detecting objects with high accuracy and speed but fall short in semantic reasoning and understanding the context, such as interpreting relationships or answering queries like "Find the object that shouldn't be here" or "Is the leash attached to the dog?" due to their limited capability to analyze beyond isolated object identification within fixed classes [121].

  - **Integration of Visual and Textual Data:** Multimodal LVLMs leverage both visual perception and language understanding to enhance detection capabilities:

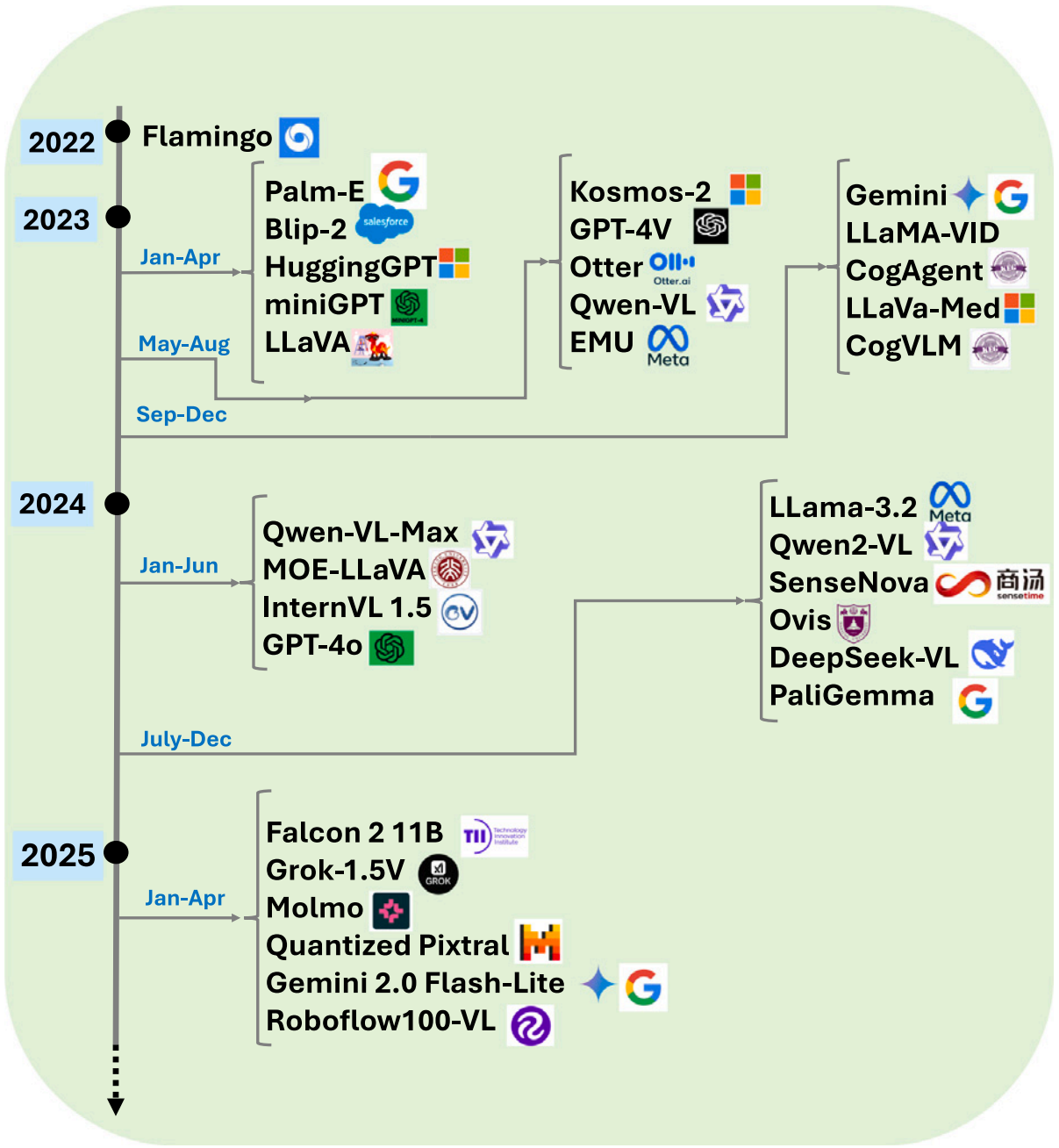

**Fig. 6.** Evolution of LVLMs for object detection. Since 2022, LVLMs have evolved from simple cross-modal encoders to highly capable generative and grounding models that support open-vocabulary detection, multimodal alignment, and real-time reasoning across diverse visual domains.

    * **Cross-Modal Attention:** These models use cross-modal attention to link specific words to corresponding image regions, enhancing detection accuracy and enabling the generation of descriptive textual content about the visual data [121].
    * **Language-Driven Visualization:** Language queries in LVLMs can directly influence the processing of visual data, beneficial in applications requiring detailed visual explanations, such as educational tools (e.g., automated grading) or advanced surveillance systems [128].
    * **Semantic Enhancement:** The integration of NLP capabilities allows LVLMs to process complex queries, such as identifying all red cars not parked next to yellow vehicles, offering a detailed understanding that extends beyond traditional object detection frameworks [128].

  - **Handling Complex and Dynamic Scenes:** LVLMs can handle complex scenarios and environments where traditional object detection models struggle.

    * **Dynamic Contextual Adaptation:** LVLMs adjust their processing based on the scene or query context, providing flexibility to effectively handle scenes of varying complexity and dynamics.
    * **Enhanced Object Recognition and Segmentation:** By integrating visual cues with contextual information from language models, LVLMs improve segmentation and recognition tasks, especially in distinguishing between objects in crowded or overlapping scenes.

**Table 2**
Recent advances in multimodal LVLMs have transformed object detection by introducing open-vocabulary reasoning and zero-shot detection and localization, driven by deep visual-textual alignment. Architectures such as DeepSeek-VL2 utilize MoE to enhance multimodal fusion; however, this table highlights LVLMs specifically tailored for object detection. It summarizes their core training datasets, spatial grounding mechanisms such as the automatic generation of bounding box tokens in Kosmos-2.5, and key architectural trade-offs. Despite these advancements, current LVLMs emphasize semantic comprehension over fine-grained localization, underscoring the need for hybrid approach with the use of conventional detectors in applications requiring precise object localization.

| LVLM name & Reference | Training data | Parameters | Tokenizer/Vision encoder | Architecture | Key features and strengths for object detection |
|---|---|---|---|---|---|
| GPT-4V [84] | Web-scale image-text (400M+) | 1.8T | CLIP-ViT-L, BPE | Decoder | Real-time processing, bounding box descriptions |
| DeepSeek-JanusPro [113] | Undisclosed | 7B | SigLIP-Large-Patch16-384 | Decoder-only | Pretrained from scratch, advanced multimodal capabilities |
| DeepSeek-VL2 [114] | WiT, WikiHow | 4.5B x 74 | SigLIP/SAMB | Decoder-Only, DeepSeekMoE | Specialized in complex query resolution and multimodal reasoning |
| Kosmos-2.5 [107] | LAION-2B+GRIT | 1.3B | ViT-L, Unigram | Enc-Dec | Zero-shot detection via spatial tokens |
| InstructBLIP [85] | CoCo, VQAv2 | 13B | ViT, Flan-T5/Vicuna | Encoder-Decoder | General-purpose vision-language model with versatile applications |
| LLaVA-Next [86] | CC3M+SBU+COCO | 7B | CLIP-ViT-L, LLaMA-2 | Decoder | Bounding box outputs for VQA |
| Grounding DINO 1.5 [115] | COCO+LVIS | 110M | Swin-B, BERT | Encoder | LLM-integrated zero-shot (47.7 mAP) |
| Florence-2 [116] | FLD-900M | 5B | ViT-g, T5 | Enc-Dec | Unified detection & captioning |
| YOLO-World [117] | Objects365+OpenImages | 42M | YOLO-CSP, CLIP | Encoder | Open-vocab real-time (60+ FPS) |
| YOLOE [118] | Diverse open prompt mechanisms (Text, Visual, and Prompt-Free) | 1.3B | RepRTA, SAVPE | Unified Encoder-Decoder | High efficiency, real-time seeing, zero-shot performance, across diverse prompts |
| Flamingo [73] | M3W ALIGN | 80B | Custom Encoder, Pretrained Chinchilla Backbone | Decoder Only | Advanced open-vocabulary reasoning, strong cross-modal alignment |
| CogVLM [71] | LAION-2B, COYO-700M | 18B | CLIP ViT-L/14, Vicuna | Encoder-Decoder | High capacity for contextual understanding, advanced vision-language integration |
| OWL-ViT v2 [119] | ALIGN-1.8B | 630M | ViT-B/16, BPE | Encoder | Vision-language transformer (47.0 mAP) |
| DINO-GPT4-V [84] | LVIS+VG | 1.2B | DINOv2, GPT-4 | Hybrid | Two-stage detection refinement |
| Shikra [108] | VG+GRIT | 3B | ViT-L, LLaMA | Decoder | Spatial Q&A with coordinates |
| VisionLLM [120] | Object365 | 13B | ViT-L, LLaMA | Decoder | Unified detection via prompts |
| Ferret [111] | GRIT+LVIS | 7B | CLIP-ViT, LLaMA | Decoder | Hybrid region-text representations |
| Qwen-VL [121] | Wukong-200M | 9.6B | ViT-L, Qwen | Enc-Dec | Multitask detection, precise coords |
| InternLM-XC [122] | MultiInstruct-1.5M | 20B | ViT-e, InternLM | Decoder | Context-aware localization |
| BLIP-2.5 [123] | VG+SBU | 1.2B | ViT-g, BERT | Enc-Dec | LLM-enhanced visual grounding |
| GLaMM [124] | SA-1B | 3B | SAM-ViT, PaLM | Decoder | SAM-like masks with LLM reasoning |
| X-LLM [125] | WebLI-10B | 12B | ViT-22B, PaLM-2 | Enc-Dec | Pixel-level attention maps |
| 4M-Det [126] | ImageNet-21K | 86M | ViT-S, BPE | Encoder | Cross-task detection, efficient design |
| PaLI-3 [127] | WebLI-5B | 17B | ViT-22B, mT5 | Enc-Dec | LLM-scale vision-language |
| ContextDET [11] | VG+GRIT | 700M | CLIP-ViT, RoBERTa | Encoder | Interactive context-based detection |
| DeepSeek-JanusPro [113] | Undisclosed | 7B | SigLIP-Large-Patch16-384 | Decoder-only | Open-vocabulary detection via Mo and High-resolution (384 px) small-object localization |
| DeepSeek-VL [114] | WiT, WikiHow | 4.5B (74 experts) | SigLIP, SAM-B | Decoder-only MoE | Multi-task detection & caption learning |

Abbreviations: Enc-Dec = Encoder-Decoder, VQA = Visual Question Answering, mAP = mean Average Precision, FPS = Frames Per Second mAP@50 and mAP@0.5:0.95 values are based on publicly reported results from each model's original paper or benchmark, primarily using COCO (cocodataset.org), LVIS (lvisdataset.org), and custom datasets.

### 2.2. Visual analysis of object detection with multimodal LVLMs

Fig. 7 demonstrates the capabilities of multimodal LVLMs in object detection across various environments. Fig. 7a particularly focuses on SpatialLM,[1] a recent and pioneering 3D LLM, which excels in processing 3D point cloud data from diverse sources such as monocular video sequences, RGBD images, and LiDAR sensors to generate structured 3D scene understandings. This model efficiently maps unstructured 3D geometric data into detailed, semantically rich scenes, identifying architectural elements like walls, doors, and windows alongside oriented object bounding boxes categorized by their semantics. These advancements highlight SpatialLM's robust spatial reasoning capabilities, positioning it as an essential tool for object detection that significantly enhances applications in autonomous navigation, embodied robotics, and detailed 3D scene analysis. Likewise, Fig. 7b illustrates the effectiveness of multimodal LVLMs in object detection, depicting prediction results from TaskCLIP (dashed blue rectangle) compared to ground truth (solid red rectangle) across various tasks. As highlighted by Chen et al. [129], TaskCLIP enhances object detection by aligning visual features with task-specific textual prompts for precision, which seeks objects suitable for specific tasks. This approach combines the advantages of LVLMs' semantic richness and a calibrated embedding space for images and texts to improve object detection outcomes. TaskCLIP employs a two-stage design: general object detection followed by task-reasoning object selection. The initial stage uses pre-trained LVLMs as the backbone, providing a robust framework for interpreting complex visual-textual data. The second stage involves a transformer-based aligner that recalibrates the embeddings to align object images with their corresponding visual attributes, often described by adjective phrases. This design addresses the challenges of traditional all-in-one models, which typically lack text supervision and suffer performance due to imbalanced and scarce training datasets. Experimental results show that TaskCLIP surpasses the DETR-based TOIST model in both accuracy and efficiency, with a notable 6.2% increase in accuracy.

This two-stage framework significantly improves both the generalizability and efficiency of object detection by harnessing the rich semantic knowledge embedded in multimodal LVLMs [129]. By decoupling general detection from task-specific reasoning, TaskCLIP offers a scalable solution for precise, context-aware object identification. Such advancements are particularly valuable in real-world scenarios that demand high-level task understanding such as assistive robotics in elderly care, context-sensitive navigation in healthcare environments, and intelligent companionship systems, where detecting objects relevant to users' intent is critical. TaskCLIP thus exemplifies how LVLMs can be effectively tailored for nuanced, task-oriented object detection challenges.

[1] https://manycore-research.github.io/SpatialLM/.

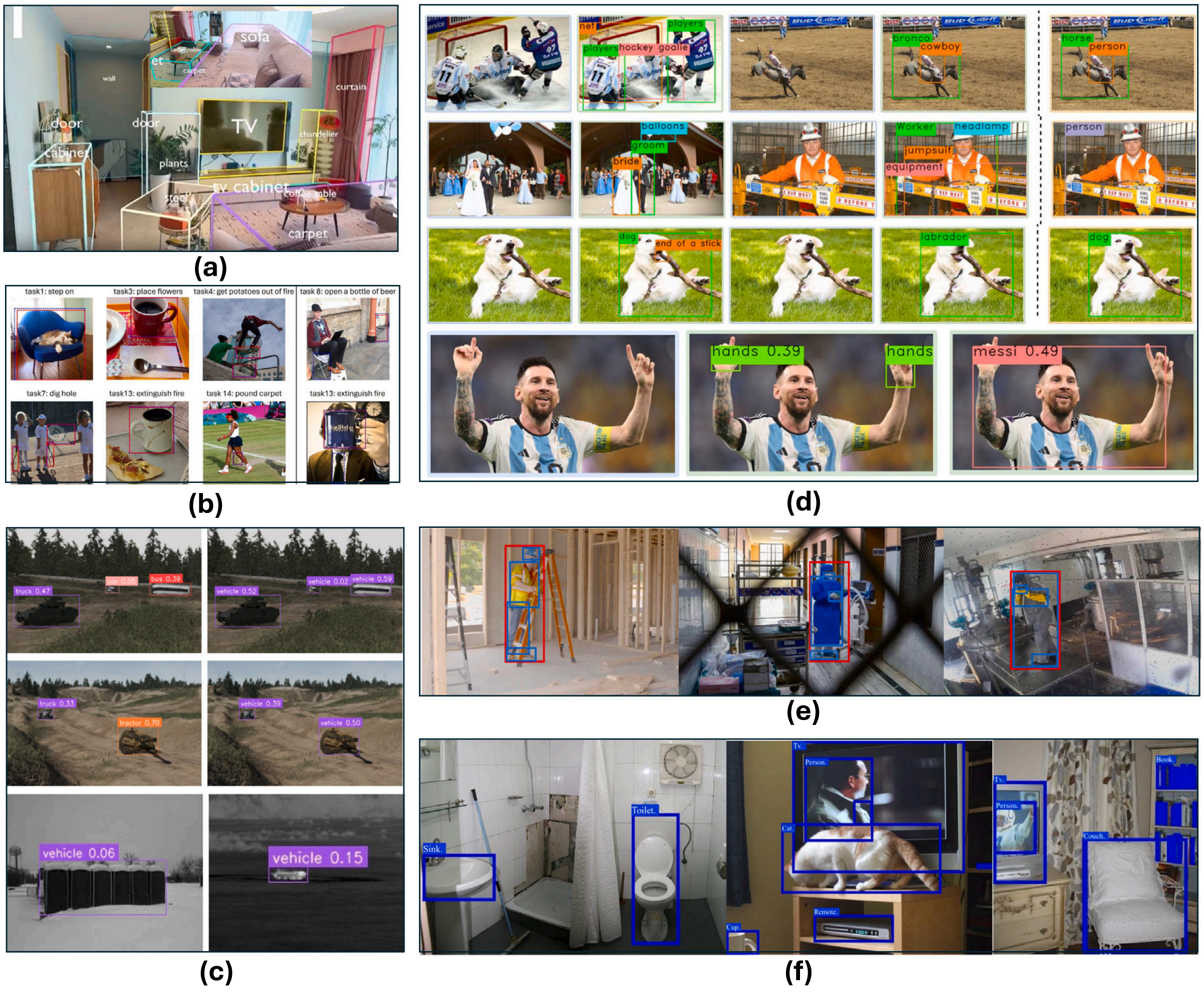

**Fig. 7.** Examples cases of object detection with LVLMs: (a) Visualization of SpatialLM's application in 3D object detection and scene understanding, demonstrating the model's ability to process point cloud data from various sources like monocular video sequences, RGBD images, and LiDAR sensors (URL: https://github.com/manycore-research/SpatialLM); (b) TaskCLIP's effectiveness in task-oriented object detection across different environments, showing both successful and unsatisfactory detection outcomes [129]; (c) Zero-Shot scene understanding for automated target recognition using LVLMs, demonstrating mis-recognition adjustments and binary detection enhancements for novel object categories [130]; (d) ContextDET implementation in contextual object detection, illustrating its ability to handle complex human-AI interaction through multimodal integration [11]; (e) Clip2Safety application in safety compliance detection within diverse workplaces, highlighting its interpretability and fine-grained detection capabilities [131]; and (f) LLMDet's open-vocabulary object detection, utilizing a LLM to enhance caption generation and detection performance across varied indoor scenes [132].

Furthermore, Fig. 7d, as explored by Zang et al. [11], presents an advanced example of object detection with multimodal LVLMs, emphasizing "Contextual Object Detection". The ContextDET model introduces a novel approach by integrating visual scenes with surrounding textual and situational context to accurately interpret and interact with objects in diverse human-AI interaction scenarios. The figure shows tasks such as completing masked object names, predicting captions with corresponding object boxes, and answering questions about object locations and names, which go beyond traditional object detection that often focuses on a limited set of predefined object classes. ContextDET innovatively addresses the gap where existing detectors fail, particularly in recognizing and localizing objects like 'hockey goalie' or 'bride' that require a clear understanding of the context. By leveraging a generate-then-detect framework, ContextDET employs a visual encoder for high-level image representation, a pre-trained LLM for text generation and multimodal context decoding, and a visual decoder to compute conditional object queries. This system not only enhances detection accuracy but also improves the model's interaction with human language, allowing for a more dynamic response to varied and specific object recognition tasks. The study by Zang et al. [11] reveals that ContextDET significantly outperforms traditional and open-vocabulary detection models in scenarios requiring detailed contextual understanding.

Moreover, Fig. 7e, as discussed in Chen et al. [131], demonstrates the Clip2Safety model's capacity for interpretable and fine-grained detection of safety compliance in diverse workplaces. This model enhances PPE detection (Personal Protective Equipment) accuracy and speed across real-world scenarios, integrating scene recognition with fine-grained verification to improve safety monitoring [131]. Lastly, Fig. 7f illustrates the application of multimodal LVLMs in diverse object detection scenarios, as explored by Fu et al. [132]. Their study introduces LLMDet, an advanced open-vocabulary detector that co-trains with a LLM to generate detailed, image-level captions, enhancing detection performance. By utilizing a specially curated dataset, GroundingCap-1M, which includes grounding labels and detailed captions for each image, LLMDet incorporates both standard grounding loss and caption generation loss in its training. This innovative approach allows LLMDet to surpass baseline models significantly, showcasing its enhanced capability to interpret and describe complex scenes accurately, thereby establishing a symbiotic enhancement of multimodal model performance [132].

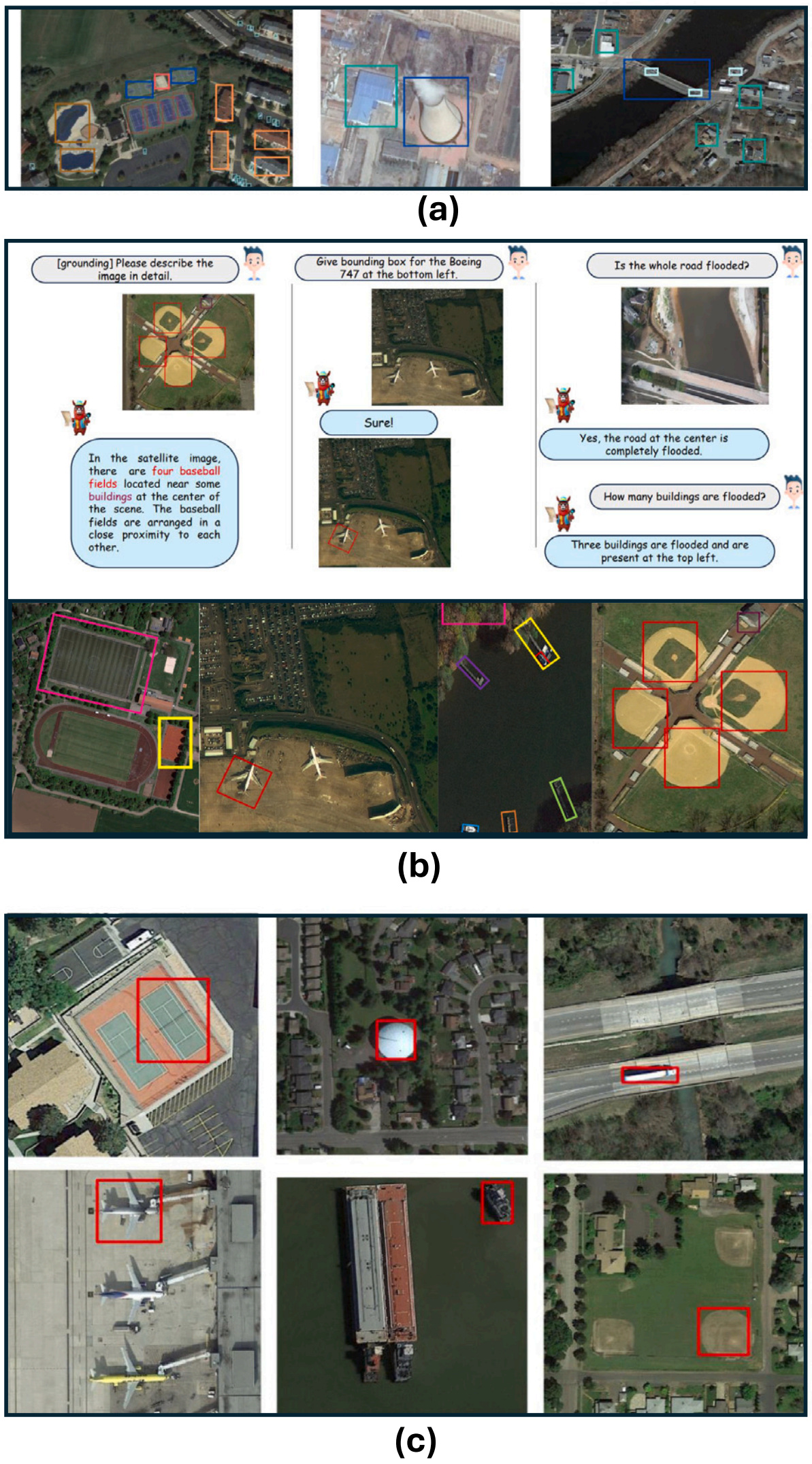

**Fig. 8.** Examples of remote sensing object detection with vision language models: (a) Visualization of advanced open-set object detection methodologies in remote sensing using multimodal LVLMs, showcasing the innovative approach of integrating LVLMs for identifying and categorizing unknown objects without manual labeling [133]; (b) Demonstrating GeoChat's capabilities in grounded, multitask conversations and robust object detection in the field of remote sensing [134]; and (c) Results of remote sensing with open vocabulary detection and scene classification by SkyEyeGPT, highlighting its enhanced performance in multi-granularity vision-language understanding tasks [135].

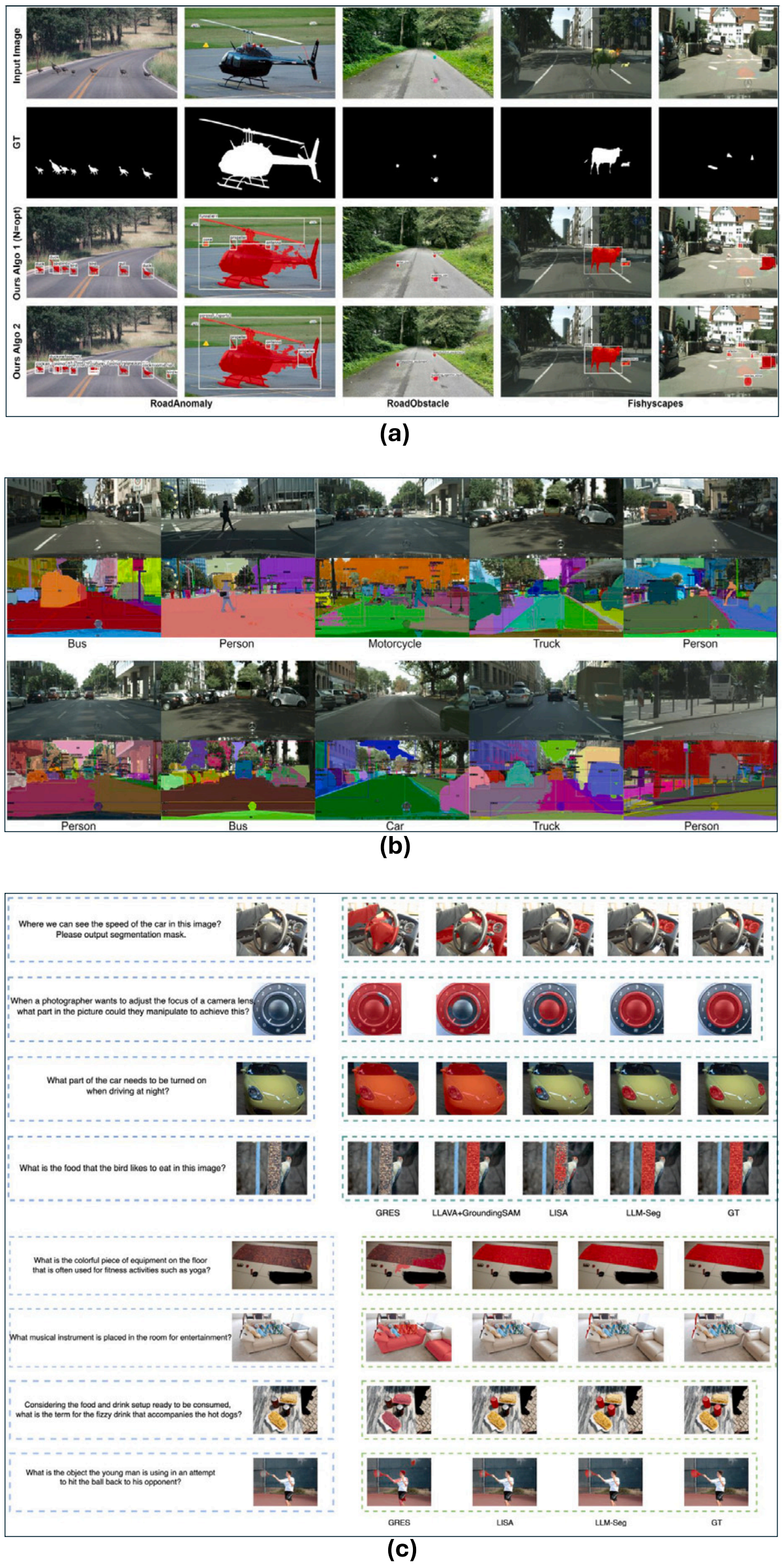

**Fig. 9.** Illustrating image segmentation with vision language models: (a) Visualization of object segmentation using multimodal LVLMs in zero-shot settings, highlighting their ability to detect and segment previously unseen (out-of-distribution) objects in complex scenes [136]; (b) Additional illustration of the robust performance of multimodal LVLMs in object segmentation, highlighting their utility in accurately detecting and segmenting objects in real-world scenarios, providing valuable insights for advancing automated perception systems [136]; and (c) Visual comparison of LLM-Seg against state-of-the-art methods, showing superior segmentation results for multiple instances and validation on the LLM-Seg40K dataset, establishing a new benchmark for reasoning segmentation approaches [137].

**Remote Sensing Object Detection with multimodal LVLMs:** Recent studies underscore significant advancement in remote sensing object detection using multimodal Language-Vision models (Fig. 8). Saini [133] developed a novel methodology for open-set object detection in remote sensing, leveraging LVLMs to identify and categorize unknown objects without manual labeling, significantly enhancing generalization over traditional methods such as YOLO and Mask R-CNN. This approach integrated advanced models to detect known objects and employs threshold-based proposals for discovering unknown categories, subsequently using LVLMs for semantic labeling, as visualized in Fig. 8a.

Additionally, Kuckreja [134] introduced GeoChat, a grounded vision-language model tailored for remote sensing, which addresses the unique challenges of high-resolution imagery and diverse object scales typical in remote sensing images. GeoChat supports multitask conversational capabilities and demonstrates robust zero-shot performance across various tasks including object detection, visually relevant conversations, and scene classification, enhancing interactivity and accuracy in remote sensing applications as shown in Fig. 8b. Furthermore, Zhan [135] developed SkyEyeGPT, a unified multimodal LLM specifically designed for remote sensing that excels in image and region-level tasks. By aligning remote sensing visual features with language domain instructions, SkyEyeGPT facilitates enhanced instruction-following and dialogue capabilities, outperforming conventional instruction-tuned LLMs such as GPT4 or LLaMa in tasks such as referring expression generation and scene classification, depicted in Fig. 8c.

**Object Segmentation with multimodal LVLMs:** In addition to enhancing object detection and localization, multimodal LVLMs have shown promising results in object segmentation, as highlighted in Fig. 9. The study employing the zPROD framework as shown in Fig. 9a demonstrates significant advancements in zero-shot, open vocabulary object detection and segmentation within automated driving contexts [133]. This novel approach targets the accurate detection and segmentation of out-of-distribution (OOD) objects on roads, effectively leveraging LVLMs for visual grounding and comprehensive contextual interpretation.

The zPROD methodology merges detection with segmentation, enabling precise identification and characterization of previously unrecognized objects in complex driving environments. The model achieves this capability by utilizing LVLMs to generate precise and contextually appropriate predictions for both known and novel object types. The approach is evaluated against traditional fully supervised methods on established benchmarks such as SMIYC [138] and Fishyscapes [139]. In these comparisons, zPROD not only outperforms standard methods in the RoadAnomaly and RoadObstacle datasets but also achieves comparable results on Fishyscapes subsets. These benchmarks are critical for evaluating the performance of object detection and segmentation models. Specifically, testing with SMIYC and Fishyscapes helps assess how well models handle anomalous objects and challenging road obstacles not present in training data, thereby measuring their generalization to new and unpredictable scenarios.

Additionally, Fig. 9b showcases object segmentation with multimodal LVLMs, as detailed by [133]. The figure displays sample images from the FS Static dataset, which includes annotated OOD objects that actually belong to the in-domain classes of the Cityscapes dataset. The zPROD model, leveraging inference on frozen LVLMs, accurately predicts instance classifications, identifying OOD objects that are misclassified due to their presence in the in-domain list. In 21 (RA) and RoadObstacle21 (RO), out-of-distribution (OOD) objects appear in varied locations within the scene. While state-of-the-art supervised methods such as Maximum Softmax Probability (MSP-based) detectors often misclassify in-domain objects as OOD due to texture variations, LVLMs like APE ((Aligning and Prompting Everything)) more accurately classify and ground them as in-domain.

Furthermore, Wang et al. [137] advance image segmentation through their development of LLM-Seg, a framework that integrates LLM reasoning to enhance perception systems by interpreting user intentions for target object segmentation [137]. As depicted in Fig. 9c, LLM-Seg outperforms state-of-the-art methods (e.g., GRES, LISA, LLaVA+Grounding SAM) in visual comparisons, particularly excelling in multiple instance scenarios. The lower rows of Fig. 9c feature results from the LLM-Seg40K validation split, demonstrating the efficacy of fine-tuned models such as LLaVA + Grounding SAM, LLM-Seg, LISA. This innovative approach establishes LLM-Seg40K as a new benchmark for reasoning segmentation, significantly contributing to the field by enabling more accurate and context-aware segmentation outcomes.

## 3. Architectural innovations and application of multimodal LVLMs for object detection

In this section, we summarize the literature findings for RQ2 as how are these models designed and implemented to enhance their object detection capabilities, and we present the architectural details and technical methodology adopted by the recent multimodal LVLMs.

### 3.1. Unified architectures and enhancement mechanisms in LVLM-based object detection

The DetGPT framework as illustrated in Fig. 10a introduces a novel reasoning-based object detection approach that combines a multimodal model with an open-vocabulary detector [16]. The multimodal model, which includes a pre-trained visual encoder and a LLM, interprets user instructions and identifies relevant objects within visual scenes. This identification process involves a cross-modal alignment where image features are mapped to the text domain using a linear projection layer. The identified objects' names or phrases are then passed to the open-vocabulary detector for precise localization in the visual space. The integration of BLIP-2 as the visual encoder and Vicuna as the language model facilitates robust interpretation and reasoning across both visual and textual features.

CoTDetas illustrated in Fig. 10b left and TaskCLIP as illustrated in Fig. 10b right side utilize two-stage frameworks to enhance task-oriented object detection by employing Large Scale Vision-Language Models [129,140]. These methodologies harness the power of pre-trained Vision-Language Models (VLMs) like CLIP and Flamingo to create high-quality, unified embeddings that align visual and textual features effectively [129]. In the initial stage, general object detection is performed while parsing the task utility into descriptive attributes using LLMs [141]. Subsequent stages involve the alignment of these attributes with visual embeddings, guided by affinity matrices generated from VLMs [142,143]. This alignment facilitates the selection of objects that fit the task requirements. Additionally, TaskCLIP introduces a transformer-based aligner that recalibrates VLM embeddings to enhance the match between visual features and specific task-related adjectives [129], thereby improving detection precision and reducing false negatives.

A notable advancement in object detection is the emergence of open-ended detection frameworks (e.g., Fig. 10c), which aim to identify and name objects without relying on predefined category sets. This capability has been possible by the architectural designs that combine region proposal networks with generative language models, enabling systems to generate object names in a free-form, context-aware manner. One such implementation is the integration of Deformable DETR with generative models, allowing for accurate region extraction alongside language-driven label generation. These models are trained end-to-end using region-word alignment loss, which ensures that the semantic content of each visual region is accurately reflected in its textual description. Such methods demonstrate strong potential in zero-shot detection settings, offering greater adaptability to novel and unseen environments. This direction is well-illustrated by recent work such as GenerateU [144], which showcases how aligning region-level visual features with language tokens significantly expands the range and flexibility of detectable object classes.

Furthermore, Zhao et al. [15] in Fig. 10d, leverages LVLMs to generate semantically relevant negative object descriptions and text-to-image diffusion models to synthesize corresponding negative images, improving upon prior rule-based or random negative sampling. To enhance model robustness, this method generates semantically related yet non-matching negative samples using instruction-tuned large language models (LLMs). These negatives help the model distinguish fine-grained differences in object descriptions. Additionally, text-to-image diffusion models (e.g., GLIGEN) generate negative images by altering bounding box content based on modified text prompts. The approach applies CLIP-based filtering to mitigate noise, ensuring only semantically valid negatives are used. This dual-synthesis process improves LVLMs' object detection accuracy and semantic understanding.

In the study by Zhou et al. (2025), the authors present an innovative approach to enhance Open Vocabulary Object Detection (OVD) by utilizing an adapter-based framework that integrates the hidden states from Multimodal LLMs (LVLMs) into the detection process [145]. This methodology diverges from traditional data generation methods that are prone to distribution shifts and overfitting, as illustrated in Fig. 11a. This method leverages the semantic richness and knowledge embedded in the early layers of LVLMs to improve the generalization capabilities of object detectors without relying on human-curated data. The core innovation of their approach lies in the introduction of a zero-initialized cross-attention adapter that effectively transfers knowledge from the LLM component of the MLLM to the object detection decoder. This adapter harnesses the intermediate hidden states from the LLM, which retain strong spatial-semantic correlations that are crucial for accurately grounding complex free-form text queries into visual representations. By doing so, the framework not only enhances the semantic richness of the detected objects but also expands the detector's ability to generalize across diverse and unseen categories. Empirical results demonstrate that this adaptation significantly boosts performance on standard benchmarks like Omnilabel, with improvements in grounding accuracy for both plain categories and complex queries. The approach also incurs a manageable increase in computational overhead, making it a practical solution for enhancing existing object detection systems.

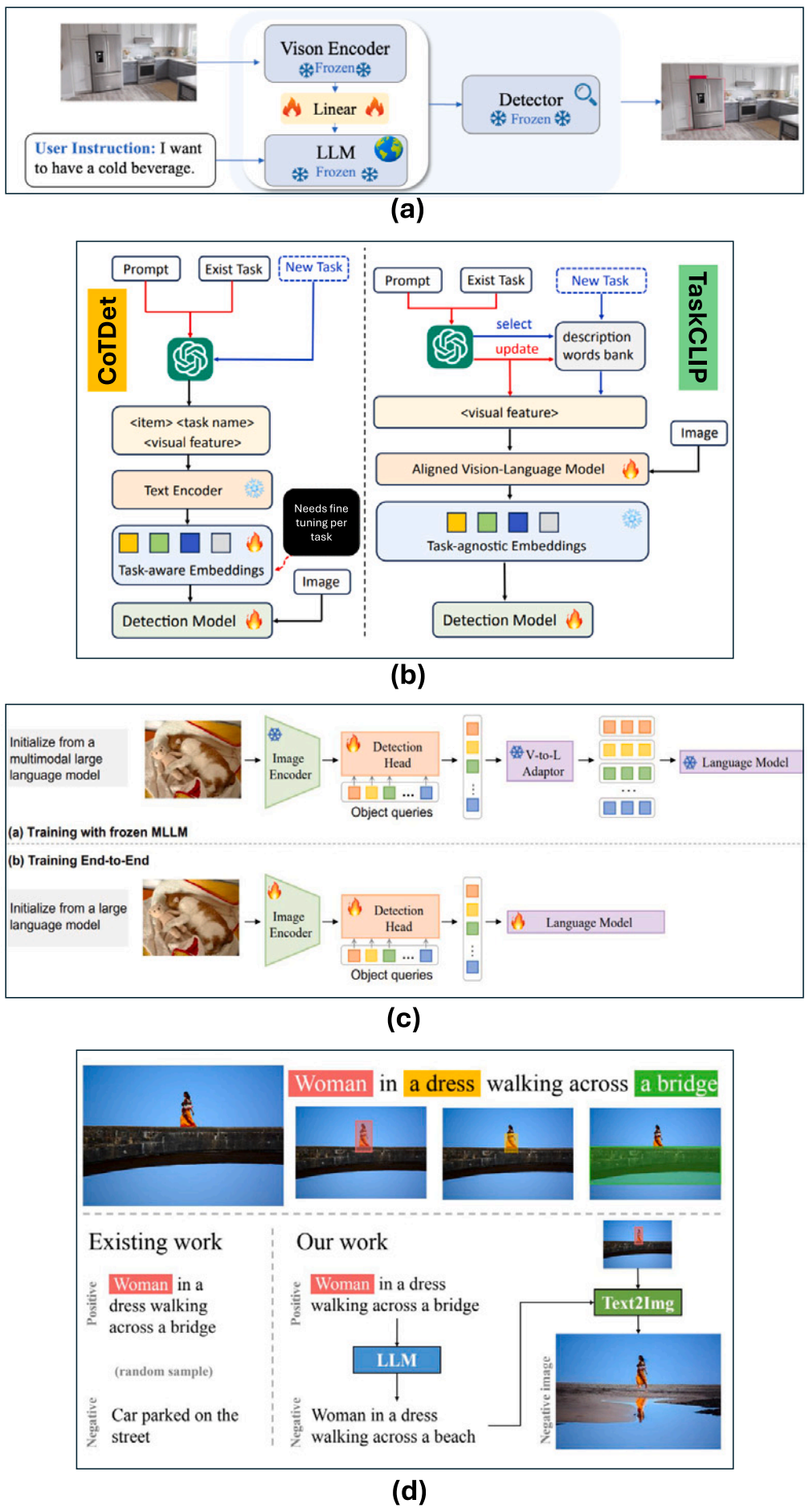

**Fig. 10.** (a) DetGPT integrates a vision encoder and LLM for user-instruction-driven detection [16]. (b) CoTDet and TaskCLIP employ vision-text aligners and grouping strategies for task-guided detection [129,140]. (c) GenerateU uses dual training for open-ended object detection [144]. (d) Zhao et al. generate negative samples for object detection using LLMs and diffusion models [15]. These LVLM-based object detection frameworks utilize a diverse architectural strategies from instruction-conditioned detection in DetGPT, task-guided alignment in CoTDet and TaskCLIP, open-ended generative detection in GenerateU, to negative sample generation using LLM-diffusion integration in Zhao et al.'s method highlighting the evolving design space for visual reasoning across modalities.

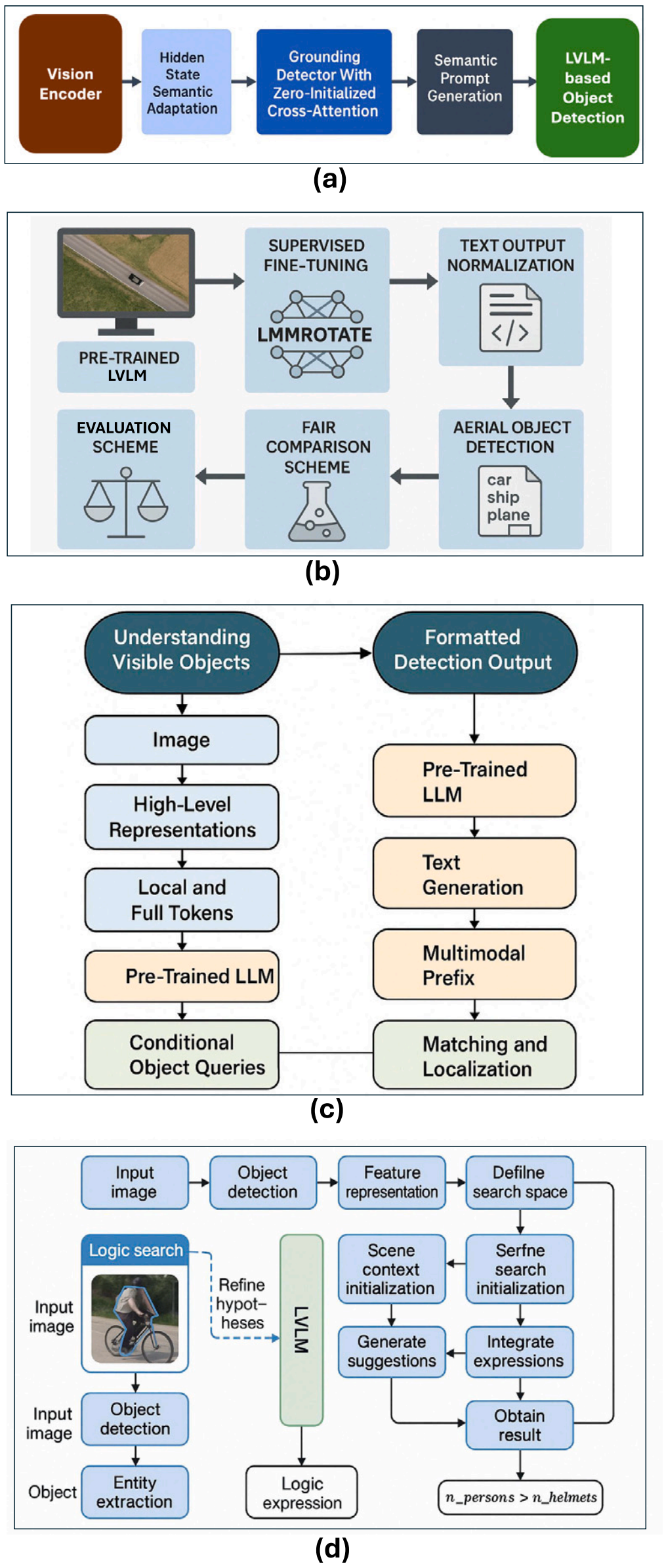

**Fig. 11.** State-of-the-art LVLM-based object detection frameworks: (a) LED aligns LLMs with visual encoders via cross-attention adapters [145]; (b) LMMRotate reformats detection outputs into textual sequences for MLMs [146]; (c) ContextDET contextualizes object detection via generate-then-detect modeling [11]; (d) VED-SR combines symbolic regression and LLMs for interpretable event detection [147].

In the study by Li et al. (2025), a novel approach is introduced for applying multimodal language models (MLMs) to the domain of object detection in aerial (or remote sensing) images, a challenge previously unexplored by RS MLMs due to the autoregressive nature of LVLMs which contrasts with the parallel output typically required for detection tasks [146]. This paper as illustrated in Fig. 11b shows a transformative method called LMMRotate, which adapts MLMs to process and output detection data by normalizing detection outputs into a textual format compatible with the MLM framework. This adaptation allows the MLM to handle object detection without altering its foundational autoregressive properties. The methodology begins with the preprocessing of remote sensing (RS) images, where the image features are extracted and flattened. These features are then projected into a tokenized space that matches the input format of the language model, facilitating the integration of visual data with textual detection instructions. This multimodal integration leverages the inherent strengths of MLMs in understanding and generating text to perform object detection tasks by translating visual inputs into descriptive text outputs. Furthermore,

LMMRotate incorporates a novel evaluation method designed to compare the performance of MLM-based detectors with traditional object detection models. This approach addresses the inherent discrepancies between the numerical output of conventional detectors and the textual output of MLMs by normalizing the detection outputs.

The evaluation strategy proposed by Li et al. [146], illustrated in Fig. 11b, ensures equitable assessment between LVLM-based and conventional detectors by excluding confidence scores and focusing on core detection components object category and polygon-based bounding box localization. Instead of adopting confidence-based mAP metrics, the study introduces the mAPnc (mean Average Precision with no confidence), which better reflects the inherent capabilities of LVLMs by eliminating reliance on additional post-processing heuristics. This normalization facilitates direct and fair comparison, showing that MLMs can deliver detection performance closely matching that of traditional models. By framing detection outputs as structured text, LMMRotate not only aligns with the generative nature of MLMs but also extends their utility into high-stakes domains such as remote sensing and aerial image interpretation.

### 3.2. Domain-specific and task-oriented multimodal detection innovations

In the study by Zang et al. (2024), ContextDET is proposed as a novel end-to-end framework for contextual object detection, utilizing multimodal LVLMs to overcome limitations of traditional object detection methods [11]. As visualized in Fig. 11c, the workflow consists of ten interlinked steps that integrate visual inputs and language cues through a generate-then-detect pipeline. Conventional models often operate on a fixed label set and struggle with open-vocabulary scenarios, but ContextDET redefines the detection process through three main tasks language Cloze Test, visual captioning, and visual question answering each contextualizing objects within human-interactive prompts. A frozen visual encoder extracts local and global representations, which, when combined with language tokens, are passed to a pre-trained LLM. The LLM generates contextual embeddings, treated as prior knowledge for the detection process. These embeddings condition object queries in a cross-attention-based visual decoder, allowing precise localization and labeling of objects described by human language. This process enables recognition of specific and relevant concepts such as ‘goalie’ or ‘groom’ instead of generic terms like ‘person’. The framework shows strong performance on the CODE benchmark, which assesses models on contextual and open-vocabulary detection tasks. By integrating flexible language-conditioned detection mechanisms, ContextDET presents a scalable and accurate approach for future AI systems interacting with complex visual environments.

In the study by Zeng et al. (2025), a novel training-free framework VED-SR (Visual Event Detection via Symbolic Regression) is proposed, marking a significant advancement in moving from traditional object detection toward comprehensive event understanding through LLM-guided symbolic reasoning [147]. As illustrated in Fig. 11d, this framework is composed of two major methodological pillars: symbolic logic search and automated reasoning with large language models (LLMs), jointly enabling a plug-and-play, interpretable, and domain-agnostic detection system. The process begins with open-set object detection, where pre-trained detectors extract structured entity-level features, such as bounding boxes, categories, and spatial relations (Steps 1–2). These are encoded into symbolic representations that serve as the input for symbolic regression (Steps 3–5), a mechanism designed to discover human-readable logical patterns that distinguish normal and anomalous events.

In addition, the symbolic reasoning pipeline introduced by Zeng et al. [147] integrates LLM-guided symbolic regression with open-set object detection to enable interpretable and training-free anomaly event detection. This approach is guided by LLMs through a structured prompt space comprised of scene initialization, chain-of-thought reasoning, and feedback integration (Steps 6–8). These prompts enable LLMs to semantically interpret the visual scene and propose new symbolic expressions, which are iteratively evaluated and evolved through a bidirectional interaction loop (Step 9). This loop ensures semantic consistency, interpretability, and convergence toward meaningful rules without any training data. The final output is a symbolic expression that captures high-level semantic patterns, enabling transparent and verifiable decisions for anomaly detection that can be easily audited by humans (Step 10). The framework’s robustness is validated across challenging benchmarks, including UCSD Ped2 and the newly introduced Helmet-Mac and Multi-Event datasets, where it consistently achieves or surpasses state-of-the-art detection performance. Remarkably, it does so while requiring less than 1% of the annotated data typically needed by supervised methods such as CNN- or transformer-based anomaly detection models.

Additionally, Fig. 12 presents an advanced LVLM-based segmentation and object detection architecture as proposed by Hossain et al. (2025) [92]. As illustrated in Fig. 12a, the authors introduce a dual-mode segmentation framework, “The Power of One”, that utilizes vision-language models trained on large-scale image-text pairs to perform zero-shot segmentation and object detection with minimal supervision. The framework operates in two modes: training-free inference and one-shot fine-tuning. In the training-free mode, given only class names and a query image, the model extracts text-to-image attention maps from a VLM and ranks them using an entropy-based metric called InfoScore to select the top-performing layers. These attention maps are then re-weighted using class-wise image-text matching scores, enabling robust segmentation without requiring any pixel-level supervision. In the one-shot mode, segmentation accuracy is further improved by fine-tuning the attention maps and text embeddings using a single annotated example per class. As detailed in the figure, this approach comprises components for prompt-based heatmap generation, entropy-driven layer selection, attention re-weighting, and convolutional CRF-based post-processing. This innovative pipeline significantly reduces reliance on large labeled datasets, demonstrating strong generalizability across VLMs and datasets. Overall, the approach exemplifies a scalable, interpretable, and efficient solution for open-vocabulary segmentation and contextual object understanding.

Likewise, in a recent study conducted by Wen et al. (2025), a novel architecture for Language-driven Zero-Shot Object Navigation (L-ZSON) is introduced, which was referred to as Vision Language model with a Tree-of-Thought Network (VLTNet) (Fig. 12b [148]). This architecture is structured into ten stages, aligning with four high-level modules, to facilitate semantic navigation in unseen environments without any task-specific training data. The process begins with Instruction Encoding, where natural language goals are parsed into actionable semantic cues. Next, the Visual Scene Understanding stage employs a pre-trained LVLM, such as GLIP, to detect objects, forming the basis of the robot’s situational awareness. These detections are passed into the Depth-aware Semantic Mapping stage, which fuses depth information and agent pose with semantic features to generate a layered 2D semantic map. This map is further refined through 3D Semantic Projection that encodes room and object relationships into a top-down view for downstream planning. A key component of the VLTNet framework is the Frontier Generation stage, which identifies the boundaries between explored and unexplored areas in the environment, enabling the model to detect potential regions for further exploration or object localization. The following stage, Tree-of-Thought (ToT) Prompting, introduces multi-agent reflective reasoning by prompting the LVLM to simulate expert thought processes. This is coupled with ’Tree Search and Evaluation’, where multiple reasoning paths are explored, scored, and pruned to yield globally optimized decisions on exploration paths. Upon reaching a potential target, a ’Goal Matching and Verification’ module assesses object attributes and spatial context using both vision and language grounding to verify goal satisfaction. In ’Contextual Comparison’, a second-level reasoning process compares the detected object context with the initial instruction for semantic alignment. Finally, Action Generation executes the navigation steps to approach or adjust based on the validated goal. This comprehensive framework leverages the common-sense reasoning capabilities of LLMs via ToT mechanisms to enhance exploration and decision-making in unstructured environments. Evaluations on the PASTURE and RoboTHOR benchmarks confirm VLTNet's superior performance in scenarios requiring complex language grounding and real-time scene understanding, thereby establishing a robust paradigm for scalable, zero-shot object detection and localization [148].

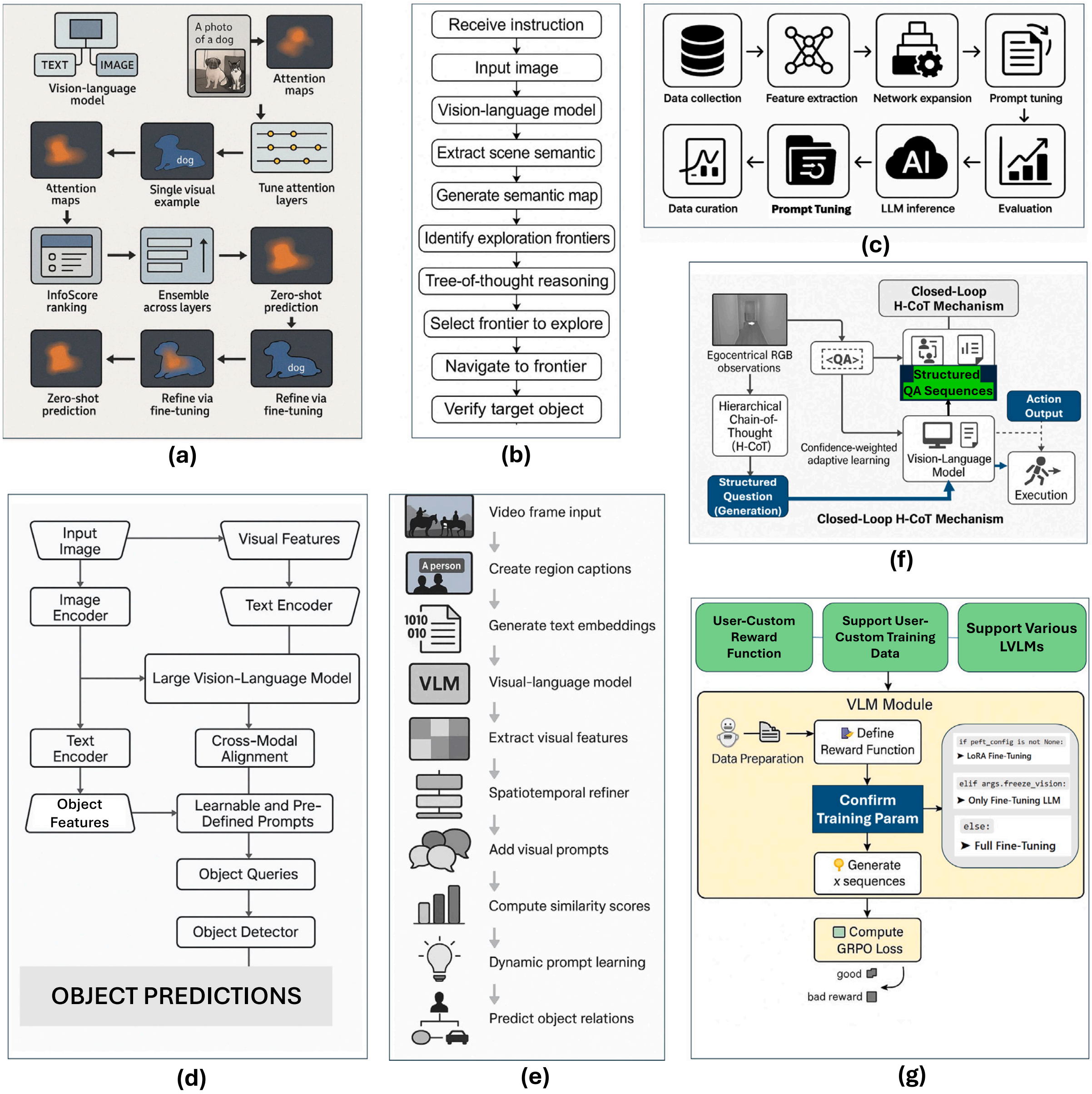

**Fig. 12.** Methodological and architectural overview of seven representative LVLM-based object detection frameworks; (a) Hossain et al. [92] propose a segmentation framework using class-specific prompts and InfoScore-guided attention for zero-shot detection without labeled data; (b) Wen et al. [148] introduce VLTNet, a Tree-of-Thought LVLM that integrates vision-language reasoning for robotic navigation; (c) Luo et al. [149] design a two-phase fire monitoring system with a FireAgent module leveraging multimodal data and LLM-guided subtasks; (d) Wang et al. [137] present LLM-Seg, combining SAM mask proposals and LLM-driven token-guided reasoning for segmentation; (e) Liu et al. [150] develop OpenVidVRD, using region captions and prompt-aligned spatiotemporal refinement for visual relation detection; (f) Cai et al. [151] propose CL-CoTNav, decomposing navigation into perception and planning via H-CoT reasoning; and (g) Shen et al. [152] introduce VLM-R1, an RL-based architecture applying GRPO to improve object detection via task-specific reward design.

Additionally, the study by Luo et al. (2025) presents a comprehensive two-phase architecture for fire event identification and impact assessment, as illustrated in Fig. 12c, which marks a significant evolution in the application of Vision-Language Models (VLMs) for environmental monitoring [149]. The proposed framework begins with the fusion of multi-band satellite imagery specifically, RED, SWIR1, and SWIR2 channels and heterogeneous environmental data, including elevation, land cover, and population density, to generate rich semantic feature maps for fire detection. This phase utilizes an enhanced object detection pipeline based on a modified YOLOv8 architecture, integrated with environmental features via weighted fusion, and regularized by the Normalized Wasserstein Distance (NWD) loss. This addition effectively improves the sensitivity to small-scale fires and enhances spatial localization performance.

In the second phase, the architecture transitions from detection to assessment through the FireAgent, an LLM-empowered decision-making module. FireAgent decomposes fire impact evaluation into a sequence of subtasks such as social sentiment extraction, rescue needs analysis, and ecological damage estimation using structured prompts and reasoning capabilities. Each subtask is handled autonomously through the agent's brain, knowledge center, and action executor, ensuring context-aware reasoning and report generation. By dynamically combining inputs from satellite imagery, social media, and environmental metadata, FireAgent synthesizes a comprehensive report detailing affected areas, fire categories, and decision-critical insights for emergency responders. This multi-step framework exemplifies how cross-domain knowledge integration and LLM-guided task planning increase the efficacy of fire event detection, tracking, and situational assessment, thereby setting a benchmark in real-time geospatial intelligence systems [149].

In the recent work by Wang et al. (2024), a novel segmentation framework named LLM-Seg is proposed, pioneering a two-stage methodology that incorporates large language model (LLM) reasoning with state-of-the-art visual segmentation capabilities [137]. As illustrated in Fig. 12d, the architecture integrates three key components: (i) an image encoder for extracting visual features from the input image; (ii) a vision-language model for fusing visual and textual information via cross-modal alignment and prompt conditioning; and (iii) an object detector that generates the final object predictions based on learned object queries. Although models such as SAM and DINOv2 are commonly used in related frameworks, they are not explicitly part of the current illustrated architecture. A valuable innovation in this architecture is the use of a special <SEG> token, which embeds segmentation intent into the LLM's input, facilitating a unified representation of user instructions and visual understanding. The segmentation process begins with SAM's "Everything Mode", which samples dense point prompts across the image to propose a wide array of potential object masks. These masks are then converted into mask embeddings using features from the image encoder. The fusion module, equipped with cross-attention and self-attention layers, aligns these embeddings with the <SEG> token. Subsequently, a dual-head mask selection module comprising an IoU head for selecting the most precise single mask and an Intersection over Prediction (IoP) head for selecting multiple relevant masks is used to compute similarity scores and regress predictions, refining the final segmentation output. A threshold-based decision mechanism ensures only the most relevant masks are retained. Notably, the architecture supports both learnable prompts and hand-crafted prompts, dynamically adapting to varied reasoning tasks. This design not only reduces the dependency on large-scale fine-tuning but also preserves generalization by freezing the core models. Furthermore, the introduction of the LLM-Seg40K dataset constructed via a GPT-4 powered data generation pipeline establishes a new benchmark for evaluating reasoning-aware segmentation. With these capabilities, LLM-Seg represents a significant advancement in vision-language integration, enabling intelligent segmentation driven by complex, human-like reasoning [137].

In a recent work by Liu et al. (2025), the OpenVidVRD framework is proposed as a transformative solution for open-vocabulary video visual relation detection (VidVRD), addressing the complexities of dynamic object interactions in temporal video streams [150]. As depicted in Fig. 12e, OpenVidVRD architecture is structured into a comprehensive ten-stage pipeline that capitalizes on the strengths of large vision-language models (LVLMs) through prompt-driven semantic space alignment. The process initiates with object trajectory extraction, where a pretrained detector captures temporally aligned bounding boxes for subjects and objects. These region proposals are fed into a region captioning module using a VQA model (e.g., BLIP-2) to generate localized descriptions for each visual region, enriching the semantic grounding. The extracted captions are encoded and fused using a visual-text aggregation module, which integrates visual and textual features across four distinct semantic roles: subject, object, union (capturing the interaction region between entities), and background (providing contextual scene information). To handle temporal dynamics, a spatiotemporal refiner module is introduced, comprising sequential spatial and temporal Transformers. This module enables the fusion of cross-modal features, augmented by motion cues and role-specific embeddings, thereby refining relational representations across frames. Importantly, OpenVidVRD introduces a prompt-driven semantic alignment mechanism that dynamically combines learnable prompts with hand-crafted ones. This hybrid prompting strategy enhances the model's adaptability to both base and novel relation categories during inference. For classification, the model computes similarity scores between refined visual features and text embeddings using CLIP, while additional adapter layers facilitate better generalization to novel concepts. This similarity-based classification mechanism is followed by open-vocabulary relation prediction, where relation prompts are decomposed and recombined with contextual embeddings to improve relational inference. Finally, the training is supervised using three objectives contrastive losses on objects and relations, and an interaction loss to guide frame-level co-occurrence modeling. Collectively, this modular and scalable architecture enables OpenVidVRD to achieve state-of-the-art results on VidVRD and VidOR benchmarks. By aligning visual semantics with language through spatial, temporal, and prompt-based reasoning, OpenVidVRD sets a new paradigm in open-vocabulary visual relation detection across diverse and unstructured video environments [150].

Cai et al. [151] introduced **CL-CoTNav**, a vision-language-based architecture designed for zero-shot Object Navigation (ObjectNav). The model integrates hierarchical chain-of-thought (H-CoT) prompting with confidence-weighted closed-loop learning to improve reasoning and adaptability. As illustrated in Fig. 12f, the framework decomposes the navigation process into two phases perception and planning using a multi-turn question-answering mechanism that enables compositional reasoning. During training, confidence scores are used to modulate the loss function, placing greater emphasis on reliable visual-textual cues. This approach significantly improves generalization to unseen scenes and novel objects, setting a new benchmark for LVLM-based object detection and decision-making in complex navigation tasks.

Additionally, in a recent study by Shen et al. (2025), a novel reinforcement learning (RL)-based framework titled VLM-R1 was introduced to enhance object detection and visual understanding capabilities in LVLMs [152]. This framework adapts the successful R1-style RL methodology from language modeling to the vision-language domain. The architecture and methodology of VLM-R1, illustrated in Fig. 12g, consists of two major components: data preparation and reward function definition (via grpo-jsonl.py), and GRPO-based RL training (grpo-trainer.py). These components work synergistically to train LVLMs through a process of sequence generation, reward computation, and policy optimization. VLM-R1 is built on the Group Relative Policy Optimization (GRPO) algorithm, which directly compares sampled responses using a reward function without the need for a separate critic model. During training, the VLM generates multiple output sequences in response to a visual-text query. These sequences are then evaluated using a carefully designed reward function, which determines their relative quality and guides model updates through GRPO loss. The framework supports flexible training paradigms such as LoRA fine-tuning, freezing of the vision tower (i.e., the image encoder backbone responsible for extracting visual features), or full-parameter optimization, making it adaptable to various computational constraints.

A major innovation of VLM-R1 is its support for custom reward functions, which were developed for two key tasks: Referring Expression Comprehension (REC) and Open-Vocabulary Object Detection (OVD). These tasks share a bounding box output format but vary in complexity. For REC, the model predicts a single bounding box from a text description, while OVD requires detection of multiple object-label pairs. The accuracy reward for REC is based on IoU between predicted and ground-truth boxes, whereas OVD rewards are based on mean Average Precision (mAP), augmented with a redundancy penalty factor (s-ovd) to prevent reward hacking. In both cases, format rewards ensure compliance with structured response expectations, enforcing JSON-style outputs within designated tags. To manage interactions across various VLM architectures, VLM-R1 introduces a modular VLM component, abstracting prompt formatting, model instantiation, and input preprocessing. This allows seamless support for models such as Qwen2.5-VL, InternVL, and LlavaNext. Experiments demonstrate that RL-trained models using VLM-R1 consistently outperform supervised fine-tuned (SFT) counterparts, especially in out-of-domain generalization. In REC, VLM-R1 achieves higher accuracy on reasoning-intensive benchmarks like LISA-Grounding [152]. In OVD, the RL model achieves 31.01 mAP on OVDEval, surpassing both SFT models and specialized detection architectures like OmDet in categories demanding semantic reasoning, such as relationship, position, and negation [153,154]. Furthermore, the study reveals key insights into reward hacking, emphasizing the importance of reward engineering. The proposed 'Length' reward mitigates excessive prediction behaviors and enables an "OD aha moment", where the model first reasons about object presence before accurate localization. Additional findings highlight the role of training data complexity and model scale in shaping RL effectiveness.

### 3.3. Major applications of multimodal LVLMs for object detection

LVLM-based object detection has revolutionized various application fields by enhancing the accuracy and efficiency of recognizing and processing visual information. For instance, in autonomous driving, LVLMs have been pivotal in improving safety and security measures. Wase et al. [12] utilized a model fusion approach, VOLTRON, integrating YOLOv8 with LLaMA2 to enhance real-time hazard identification, significantly improving object detection accuracy in dynamic driving environments. This innovation is crucial for developing autonomous vehicles that can reliably navigate complex traffic scenarios. In surveillance systems, the capability of LVLMs to parse complex scenes has been utilized to enhance security monitoring. Xie et al. [14] introduced a language-guided detection framework, which employs dynamic alignment modules to process multi-stage descriptions, which improves the surveillance system's ability to monitor and manage urban environments effectively. This application demonstrates how LVLMs can be adapted to maintain safety and order in public spaces. Furthermore, significant advancements has been observed in Robotics with the integration of LVLMs. Pi et al. [16] developed DetGPT, a reasoning-based detection model that enhances human-AI interaction within robotic systems, facilitating more effective autonomous navigation and steering, and query-based searches. This development shows the potential of LVLMs to create more interactive and autonomous robotic systems that can perform complex tasks with minimal human intervention. Remote sensing has also benefited from LVLMs, especially in the identification and categorization of unknown objects. Saini et al. [133] leveraged a multimodal approach that utilizes both satellite imagery and textual annotations to improve the detection and monitoring of environmental changes and anomalies. In low-resource scenarios, Zhou et al. [145] demonstrated how LVLMs could enhance object detection without the extensive need for curated data. By using latent semantic transfer and cross-attention adaptation, their model showed improved performance on challenging benchmarks, which is crucial for applications in regions with limited technological infrastructure. These examples illustrate the transformative impact of LVLMs across diverse sectors, driving innovations that leverage multimodal data to enhance object detection and interaction capabilities. A detailed analysis of recent advancements in LVLMs, focusing on their applications, innovations, and technical approaches, is presented in Table 3, providing a comprehensive overview of how these models are being employed to advance various fields.

## 4. Comparison between multimodal LVLMs and traditional deep learning: Capabilities and limitations

### 4.1. Performance metrics and real-time capabilities of LVLMs for object detection:

The emergence of LVLMs has drastically altered the landscape of object detection, offering distinct advantages over traditional deep learning methods such as YOLO, SSD, and Faster R-CNN. Unlike these traditional deep learning models that rely on a fixed set of detectable objects, LVLMs utilize advanced vision transformers and language models to facilitate dynamic, open-vocabulary detection. This capability enables them to interpret and respond to a broader array of objects and environments, often surpassing traditional methods in versatility and contextual understanding. For instance, Google's OWL-ViT, as cited in Zang et al. (2024) [11], shows superior performance on zero-shot detection tasks, demonstrating a significant leap in how machines understand visual content through language. Similarly, the VOLTRON model integrates YOLOv8 with LLaMA2 to enhance detection in safety-critical applications such as autonomous driving [12], emphasizing the potential of LVLMs in real-world applications that require immediate and accurate object recognition.

However, these capabilities come with trade-offs, primarily concerning computational efficiency and operational speed. Traditional models like YOLO and Faster R-CNN achieve substantially higher computational speed in object detection thus making them suitable for real-time applications and for applications with limited computational resources. On the other hand, LVLMs often require extensive computing resources, making them less ideal for these applications or environments [16,145].

The training paradigms between these types of object detection systems also differ significantly. LVLMs require extensive, diverse datasets and substantial computational resources for training, reflecting a stark contrast to the more streamlined, less resource-intensive training requirements of traditional deep learning models. However, the integration of language models enables LVLMs to perform more complex reasoning tasks, adding layers of contextual understanding that traditional systems typically lack.

These differences are critically analyzed and summarized in Table 4. This table details how each system performs across various metrics such as inference speed, accuracy, and application suitability, offering a clear view of where each technology excels or falls short.

Recent studies in LVLM-based detection systems further advance existing performance baselines by expanding resolution, task unification, and self-alignment strategies. The Griffon family of models Griffon [163], Griffon v2 [164], and Griffon-G [163] demonstrates that object localization can be effectively achieved at any spatial granularity via prompt-conditioned token-level alignment. Griffon v2 scales this approach of token-level grounding and multi-scale vision-language alignment to high-resolution visual encoders, enabling dense object prediction and accurate understanding of complex, descriptive text inputs. Griffon-G bridges detection, segmentation, and visual grounding within a shared multimodal framework, reducing architectural fragmentation and improving performance on ODinW and RefCOCO benchmarks [163,165]. These models emphasize the need for more dynamic evaluation metrics that account for spatial-textual co-reference accuracy, task transferability, and resolution-aware reasoning.

Another emerging framework is vision-guided reinforcement learning for VLM alignment, exemplified by Vision-R1 [165]. Instead of relying on human instruction tuning or supervised data captioning, Vision-R1 learns optimal vision-language mappings through an iterative reward-based curriculum [166,167]. This reinforcement-driven optimization process allows the model to autonomously learn vision-language correspondences by maximizing task-specific rewards, thereby enabling object detection, image captioning, and phrase grounding without manual annotations making it well-suited for scalable, annotation-free deployment in open-world environments. Performance-wise, Vision-R1 outperforms fine-tuned counterparts on VQAv2 (Source Link), RefCOCOg (Source Link), and LVIS detection tasks while maintaining strong zero-shot compositional generalization [168]. These contributions highlight a shift toward more autonomous, resolution-scalable, and unified architectures, redefining what constitutes efficiency and real-time viability in next-generation LVLMs.

**Table 3**
Summary of advancements in object detection using LVLMs across diverse application areas. The table first presents the title, reference of each study, and their key innovations and the specific technical approaches employed. Then, a brief description has been added summarize the practical implications and operational environments where these innovations are applied.

| Title and Reference | Innovation | Technical approach | Application context |
|---|---|---|---|
| 1. "Contextual Object Detection with Multimodal LLMs" [11] | **ContextDET**: Novel generate-then-detect framework | - Multimodal context modeling with LLM<br>- Visual encoder for high-level image representations<br>- Visual decoder for bounding boxes from language inputs | Human-AI interaction:<br>• Language-driven object detection<br>• CODE benchmark for open-vocabulary detection<br>• Extensive application in dynamic contextual settings |
| 2. Object detection meets LLMS: model fusion for safety and security [12] | **VOLTRON**: YOLOv8-LLaMA2 integration | - Single-layer architecture fusion<br>- Probability-to-text conversion<br>- LoRA optimization (7B params) | Self-driving vehicles:<br>• Small object detection (≥88% acc.)<br>• Real-time hazard identification |
| 3. Llms meet vlms: Boost open vocabulary object detection with fine-grained descriptors [13] | **DVDet**: VLM-LLM synergy | - Contextual prompt conditioning<br>- Hierarchical descriptor generation<br>- CLIP-GPT3 hybrid training | General object detection:<br>• COCO (+3.4 AP)<br>• LVIS benchmarks<br>• Rare category handling |
| 4. Described object detection: Liberating object detection with flexible expressions [14] | **DOD Framework**: Language-guided detection | - Vision-language pre-training<br>- Dynamic alignment module<br>- Multi-stage description processing | Surveillance systems:<br>• Complex scene parsing<br>• Security monitoring<br>• Urban management |
| 5. Generating Enhanced Negatives for Training Language-Based Object Detectors [15] | Synthetic negative generation | - LLM-based negative sampling<br>- Diffusion model integration<br>- Hard example mining | Model robustness:<br>• Reduced false positives<br>• Challenging benchmark handling<br>• Cross-domain adaptation |
| 6. DetGPT: Detect What You Need via Reasoning[16] | **DetGPT**: Reasoning-based detection | - Multimodal encoder-decoder<br>- Instruction tuning framework<br>- Open-vocabulary adapter | Human-AI interaction:<br>• Robotic systems<br>• Autonomous driving<br>• Query-based search |
| 7. LED: LLM Enhanced Open-Vocabulary Object Detection without Human Curated Data Generation [145] | **LED**: Zero-curated detection | - Latent semantic transfer<br>- Cross-attention adaptation<br>- Multimodal pretraining | Low-resource environments:<br>• RefCOCO (+5.2%)<br>• OmniLabel benchmarks<br>• Bias reduction |
| 8. "Advancing Open-Set Object Detection in Remote Sensing Using Multimodal LLMs" [133] | **Open-set object detection with LVLMs**: Using threshold-based region proposals and MLLM textual annotation | - Dual approach with region detection and MLLM-based discovery<br>- Integration of DOTA, DIOR, and NWPU VHR10 datasets<br>- Use of vision-language similarity metrics for validation | Remote sensing:<br>• Identification and categorization of unknown objects<br>• Significant improvement in detection and discovery metrics<br>• Enhanced generalization of models to real-world open-set conditions |
| 9. "LLMDet: Learning Strong Open-Vocabulary Object Detectors under the Supervision of LLMs" [132] | **Enhancing open-vocabulary object detection**: Integrates image-level captioning with detection training | - Utilizes GroundingCap-1M dataset with image-level captions<br>- Employs both standard grounding and caption generation loss<br>- Leverages LLM for detailed caption generation | Remote sensing:<br>• Superior open-vocabulary performance with detailed language-based supervision<br>• Demonstrates effective transfer learning capabilities<br>• Provides groundwork for stronger multimodal model integration |
| 10. "Visual LLMs for Generalized and Specialized Object Detection Tasks" [154] | **Advances in visual-language integration**: Enhances the capabilities of visual-language models by leveraging the reasoning and multitasking strengths of LLMs (LLMs) | - Discusses the evolution from conventional VLMs to highly capable VLLMs<br>- Focuses on unified embeddings for enhanced multi-task and reasoning abilities<br>- Examines specialized applications across diverse modalities | General and specialized applications:<br>• Provides a comprehensive view of VLLMs' potential in diverse scenarios<br>• Highlights the integration of advanced LLM features into visual-language tasks<br>• Paves the way for future innovations in multimodal AI systems |
| 11. "TaskCLIP: Extend Large Vision-Language Model for Task Oriented Object Detection" [129] | **Natural two-stage design with enhanced task reasoning**: Utilizes VLMs for robust semantic knowledge and aligning object detection with task requirements. | - Employs a transformer-based aligner to recalibrate VLM embeddings for accurate task-oriented object selection<br>- Incorporates a trainable score function to refine VLM matching results, improving selection precision | Task-oriented object detection:<br>• Outperforms traditional models in accuracy and efficiency on COCO-Tasks dataset<br>• Demonstrates improved generalizability and application in real-world scenarios where task requirements are complex and varied |

(*continued on next page*)

**Table 3** (*continued*).

| | | | |
|---|---|---|---|
| 12. “Enhancing Object Detection by Leveraging LLMs for Contextual Knowledge” [64] | **Contextual enhancement of object detection**: Utilizes LLaMA to improve detection in visually challenging scenarios by incorporating contextual understanding akin to human perception. | - Integrates YOLO with LLaMA to utilize high-confidence object detections<br>- Employs contextual knowledge from LLMs to predict object presence, enhancing detection accuracy under adverse conditions like occlusion | Object detection under challenging conditions:<br>• Demonstrates significant improvements in detection accuracy, especially in adverse conditions like fog and occlusion<br>• Shows the robustness of combining traditional object detection models with LLMs for contextual reasoning |
| 13. “Generative Region-Language Pretraining for Open-Ended Object Detection” [144] | **Advancing open-ended object detection**: Introduces GenerateU for generative object detection without predefined categories, using Deformable DETR and language models for region-to-name translation. | - Employs Deformable DETR for region proposal and pairs with a language model to translate visual regions into object names<br>- Utilizes a generative approach to formulating object detection, enabling the model to operate without predefined categories | Open-ended object detection:<br>• Allows detection of objects without prior categorical knowledge, enhancing flexibility and applicability in dynamic environments<br>• Demonstrates robust zero-shot detection performance on the LVIS dataset, showcasing potential for real-world application |
| 14. “Vision Language Model for Interpretable and Fine-Grained Detection of Safety Compliance in Diverse Workplaces” [131] | **Clip2Safety**: Enhanced safety compliance detection | - Scene recognition for scenario-based gear identification<br>- Visual prompts for cue generation<br>- Safety gear detection to verify compliance | Workplace Safety:<br>• Implements PPE compliance checks across diverse environments<br>• Integrates visual and language cues for enhanced detection accuracy<br>• Demonstrates significant improvements in speed and accuracy compared to traditional models |
| 15. “SkyEyeGPT: Unifying Remote Sensing Vision-Language Tasks via Instruction Tuning with Large Language Model” [135] | **SkyEyeGPT**: Integration of RS Vision-Language Tasks | - Unified vision-language model for remote sensing<br>- Aligns RS visual features with language domain via an alignment layer<br>- Employs a two-stage tuning method to enhance multi-granularity instruction-following | Remote sensing:<br>• Applies to multi-granularity vision-language tasks across 8 datasets<br>• Demonstrates superior performance in tasks like captioning and visual grounding<br>• Provides a robust dataset and tools for advancing RS-MLLM applications |
| 16. “LLMFormer: LLM for Open-Vocabulary Semantic Segmentation” [155] | **LLMFormer**: Novel use of LLMs for semantic segmentation | - Utilizes LLM priors for object, attribute, and relation knowledge<br>- Introduces three novel attention modules: semantic, scaled visual, and relation attentions<br>- Enhances OV segmentation through rich LLM-based knowledge integration | Semantic segmentation:<br>• Applies to ADE20K and Pascal Context benchmarks<br>• Achieves significant improvements over state-of-the-art models such as Mask2Former, SegFormer, and OpenSeg<br>• Capable of performing segmentation without predefined classes, suitable for real-world applications |
| 17. “VisionLLM v2: An End-to-End Generalist Multimodal LLM for Hundreds of Vision-Language Tasks” [156] | **VisionLLM v2**: Generalist multimodal LLM | - Integrates ’super link’ for flexible information and gradient transmission between MLLM and task-specific decoders<br>- Employs routing tokens and super-link queries for task-specific information processing<br>- Multistage joint training on diverse vision and vision-language tasks | Multimodal Vision-Language tasks:<br>• Supports a wide array of tasks including VQA, object localization, pose estimation, and image generation<br>• Demonstrates adaptability across domains like remote sensing, and medical imaging<br>• Achieves performance comparable to specialized models on various benchmarks |
| 18. “GeoChat: Grounded Large Vision-Language Model for Remote Sensing” [134] | **GeoChat**: Remote Sensing Multitask Conversational VLM | - Integrates conversational capabilities with high-resolution RS imagery<br>- Utilizes task-specific tokens and spatial location representations for accurate region-level reasoning<br>- Employs a novel RS multimodal instruction-following dataset for diverse RS tasks | Remote sensing:<br>• Facilitates robust zero-shot performance in RS tasks like VQA, scene classification, and object detection<br>• Enhances interaction through image and region-level dialogues<br>• Sets a new benchmark for RS multimodal conversations and assessments |
| 19. “RoboLLM: Robotic Vision Tasks Grounded on Multimodal LLMs” [157] | **RoboLLM**: Generalized Framework for Robotic Vision | - Utilizes LVLMs for a unified vision framework<br>- Employs BEiT-3 backbone for enhancing task adaptability<br>- Addresses all key vision tasks in the ARMBench dataset | Robotic vision:<br>• Streamlines integration across multiple robotic vision tasks<br>• Demonstrates superior performance and efficiency in robotic manipulation scenarios<br>• Shows resilience to out-of-distribution examples, enhancing reliability in dynamic environments |

(*continued on next page*)

**Table 3** (*continued*).

| | | | |
|---|---|---|---|
| 20. "AnomalyGPT: Detecting Industrial Anomalies Using Large Vision-Language Models" [158] | **AnomalyGPT**: Novel IAD Approach | - Utilizes LVLMs for Industrial Anomaly Detection (IAD)<br>- Employs an image decoder for detailed semantic analysis<br>- Incorporates a prompt learner to fine-tune the model via embeddings | Industrial anomaly detection:<br>• Directly assesses anomalies without manual threshold setting<br>• Supports multi-turn dialogues and few-shot learning<br>• Demonstrates high accuracy and AUC on MVTec-AD dataset |
| 21. 10. "DriveVLM: The Convergence of Autonomous Driving and Large Vision-Language Models" [159] | **DriveVLM and DriveVLM-Dual**: Hybrid Autonomous System | - Integrates VLMs with traditional autonomous driving technologies<br>- Combines scene analysis, description, and hierarchical planning<br>- Proposes DriveVLM-Dual for spatial reasoning and real-time planning | Autonomous driving:<br>• Handles complex urban driving scenarios<br>• Demonstrates efficacy on nuScenes and SUP-AD datasets<br>• Deployed in real-world production vehicles |
| 22. "VLM-PL: Advanced Pseudo Labeling Approach for Class Incremental Object Detection via Vision-Language Model" [160] | **VLM-PL**: Enhancing CIOD with VLM | - Utilizes VLMs for accurate pseudo-label verification<br>- Employs prompt tuning to refine incremental learning<br>- Integrates pseudo and real ground truths effectively | Class incremental object detection:<br>• Tackles multi-scenario incremental learning<br>• Exhibits state-of-the-art performance on Pascal VOC and MS COCO<br>• Reduces model retraining needs and memory requirements |

**Table 4**
Comparison between traditional deep learning vs. multimodal large Vision language-based object detection: This table presents a summary of the strengths and weaknesses of both traditional deep learning approaches like YOLO and Mask R-CNN, and advanced LVLMs. Specifically, it highlights key distinctions in architecture, model size, and input modalities, emphasizing LVLMs' superior capacity for processing complex, multimodal inputs and producing enriched contextual outputs. The comparison shows critical trade-offs between the two categories of models, particularly in computational requirements and inference speeds, illustrating the evolving landscape where LVLMs enhance traditional methods with their robust contextual understanding and open-vocabulary capabilities, suggesting a hybrid future for comprehensive and intelligent object detection.

| Aspect | Traditional deep learning (such as YOLO, Mask R-CNN ) | Multimodal LLM-based detection |
|---|---|---|
| Architecture | CNN-based with specialized detection heads (SSD, RPN, ROI pooling [45,109,161]) | Vision-language transformers with cross-modal attention leveraging technologies from ContextDET and VOLTRON [11,12] |
| Model size | Compact (YOLOv8: 11M params[a], Mask R-CNN: 44M [8]) | Massive, incorporating models like LLaVA-1.5 [86] and BLIP-2 [123] with up to 13B parameters [15,111] |
| Input modality | Single image input | Multimodal input (image + text prompts/instructions), as utilized in OWL-ViT [119] and InstructBLIP [11,154] |
| Output type | Bounding boxes/masks with class probabilities | Bounding boxes with enriched natural language descriptions and reasoning capabilities highlighted in models like ContextDET and LED [11,145] |
| Key metrics | • mAP@0.5:0.95 (YOLOv8: 50.2% for YOLOv8m and 53.9% for YOLOv8x, Mask R-CNN: 37.1% to 38.2% - COCO dataset<br>• FPS (YOLO-NAS[b]: 450, RT-DETR: 108)<br>• IoU, Precision | • Language-guided mAP as demonstrated in DVDet and LLaVA-1.5 [13]<br>• VQA Accuracy highlighted in TaskCLIP [129]<br>• Cross-modal Retrieval Score, pertinent in the works like SkyEyeGPT [135] |
| Inference speed | Real-time performance:<br>• YOLOv8 S: (100 FPS on an NVIDIA V100 GPU with a 640 × 640 input size)<br>• Mask R-CNN: 5 FPS<br>• YOLO-NAS S: 311 FPS and YOLO-NAS M: 170 FPS | Limited speed, with advancements from models like LLaVA-1.5 and GPT-4V [111,154] |
| Training data | Curated detection datasets:<br>• COCO (118k images)<br>• Pascal VOC (11k images) | Web-scale multimodal data as utilized in models like LLaVA-1.5 and BLIP-2, with data sources like LAION-5B [162] and CC12M [15,111] |
| Hardware requirements | Edge-deployable:<br>• Jetson Orin: 30 FPS<br>• Mobile NPUs supported | Requires server-grade GPUs:<br>• 16–80 GB VRAM needed<br>• No edge deployment |
| Strengths | • Predictable latency<br>• Hardware optimization<br>• Battle-tested reliability | • Open-vocabulary detection and contextual reasoning capabilities exemplified in ContextDET and LED [11,145]<br>• Zero-shot generalization as seen in OWL-ViT and TaskCLIP [129] |
| Weaknesses | • Closed vocabulary limit<br>• No semantic understanding<br>• Manual threshold tuning | • High computational costs and complex prompt engineering as noted in DetGPT and LED [16,145]<br>• Hallucination risks in complex scenarios [132] |

[a] https://github.com/ultralytics/ultralytics.
[b] https://docs.ultralytics.com/models/yolo-nas/.

### *4.2. Adaptability and open-vocabulary capabilities of LVLMs for object detection:*

The adaptability and open-vocabulary capabilities of LVLMs represent a major shift in the way object detection is achieved. Traditional deep learning models like YOLO and Mask R-CNN are constrained by fixed vocabularies determined during their training phase, which limits their ability to detect objects outside of these predefined categories. In contrast, LVLMs excel in open-vocabulary detection, allowing them to identify a broader range of objects based on textual descriptions, even those not seen during training. For instance, the ContextDET model developed by Zang et al. [11] exemplifies how multimodal context modeling can enhance object detection, particularly in human-AI interaction scenarios where dynamic contextual understanding is crucial. Similarly, Zhou et al.'s LED model [145] leverages latent semantic transfer to improve detection in low-resource scenarios, demonstrating substantial bias reduction and performance improvement in benchmarks like RefCOCO.

The ability of LVLMs to process and understand text alongside visual data allows them to perform tasks that are impossible for traditional models. For example, the TaskCLIP model by Chen et al. [129] utilizes robust semantic knowledge to align object detection with complex task requirements, significantly outperforming traditional methods on the COCO-Tasks dataset. This capability is underpinned by their extensive pretraining on diverse datasets like LAION-5B, as used by models such as LLaVA-1.5 [111], which provide a rich foundation for understanding and generating dynamic textual prompts that guide detection.

Furthermore, models like SkyEyeGPT [135] integrate vision-language tasks within remote sensing, employing multimodal learning to excel in tasks that require high levels of domain adaptation. The generative capabilities of models like GenerateU, developed by Lin et al. [144], also illustrate the advancement in handling open-ended object detection without predefined categories, enabling flexible application in dynamically changing environments. Another important development is the DetGPT by Pi et al. [16], which introduces reasoning-based detection mechanisms that leverage multimodal encoder–decoders to improve open-vocabulary detection, showing how LVLMs can adapt to new and unforeseen objects through reasoning and contextual interpretation.

Despite these advances, the deployment of LVLMs, as summarized in Table 4, highlights a trade-off between computational efficiency and adaptive performance. While LVLMs demonstrate high flexibility and depth in understanding, their resource-intensive nature and slower inference speeds compared to traditional models like YOLO-NAS [131] pose challenges for real-time applications. However, the continuous evolution of these models suggests that future iterations may soon overcome these limitations, further enhancing the role of LVLMs in transforming object detection across varied environments.

### 4.3. System complexity and implementation challenges of LVLMs for object detection

The deployment of LVLMs in object detection faces significant barriers due to their architectural complexity and high computational demands. Unlike compact models like YOLOv8[2] and Mask R-CNN [8], LVLMs (e.g., LLaVA-1.5, GPT-4V [111]) require extensive resources, limiting edge deployment.

Furthermore, the implementation of LVLMs involves intricate integration of language and vision modalities. This integration is not only computationally intensive but also complex in terms of data alignment and synchronization between modalities. For example, the Ferret model utilizes a hybrid architecture combining vision transformers with lightweight CNNs to manage latency but still faces challenges in balancing accuracy with processing speed [111,112]. Training LVLMs also presents substantial challenges due to their reliance on large-scale multimodal datasets, such as LAION-5B or the GroundingCap-1M dataset, which are significantly larger and more varied compared to traditional image-only datasets like COCO or Pascal VOC [132]. The requirement for vast and diverse training data increases the training duration and complexity, often necessitating thousands of GPU hours, as seen with models like InstructBLIP [144]. The prompt engineering required for effective LVLM deployment further complicates their use. Designing effective prompts that can guide the detection process without leading to context misinterpretation or hallucinations requires deep understanding of both the model's language capabilities and the task-specific requirements [131].

Moreover, the inherent complexity of these models often leads to difficulties in fine-tuning, where slight modifications in parameters or training data can lead to significantly different outcomes. This sensitivity makes robust and consistent model performance a challenging goal, particularly in dynamic real-world applications where adaptability is crucial [135]. Despite these challenges, the advanced capabilities of LVLMs, such as open-vocabulary detection and contextual reasoning, provide important benefits over traditional methods. Therefore, addressing the inherent challenges of the LVLMs is crucial for exploiting these benefits through wider adoption and optimization of LVLMs in practical object detection environments, pushing the boundaries of what is possible with AI in visual understanding tasks.

### 4.4. Architectural trade-offs and computational considerations

Despite the remarkable flexibility and semantic reasoning capabilities offered by LVLMs, their architectural complexity poses a trade-off between expressive multimodal understanding and computational efficiency, particularly in terms of inference speed, memory usage, and deployment feasibility. Unlike conventional detectors like YOLOv5 (Source Link), Faster R-CNN [169], or RetinaNet [9], which are typically optimized for bounded class vocabularies, LVLMs integrate multimodal encoders, cross-attention fusion modules, and LLM backbones often ranging from 1B to over 70B parameters. For example, models such as Flamingo-80B [73] and GPT-4V [170] require dense visual tokenization and autoregressive decoding, resulting in high memory consumption and inference latency. Even moderately sized LVLMs like BLIP-2 (13B) [123] or DetGPT (7B) [16] rely on frozen LLMs and external vision backbones, introducing multi-stage bottlenecks, where visual features must first be encoded and then passed through additional fusion layers before language reasoning, leading to increased latency, memory overhead, and integration complexity during both training and inference. These designs, while powerful for tasks such as grounding arbitrary phrases or generating spatial reasoning descriptions, are computationally limiting for real-time deployment on edge devices or embedded systems without aggressive quantization or pruning strategies.

Moreover, the performance gains offered by LVLMs often appear marginal when their computational costs are considered. On standard detection tasks, the difference between LVLM-based models and efficient transformers like RT-DETR [39] or YOLOv8X (Source Link) may within a few mAP points. For instance, GLIP scores 49.8 mAP on COCO Zero-Shot while OV-DETR reports 38.2 mAP on COCO novel classes yet both models require 4× −10× more computational resources than traditional models with comparable detection heads [171]. While models like Grounding DINO incorporate strong visual grounding and transformer refinements [41], their inference FPS is typically in the range of 30–45 on high-end GPUs, which is significantly lower than the 110–161 FPS achieved by optimized YOLO derivatives (e.g., YOLO-World [117], YOLOE [118]) on similar hardware. This discrepancy becomes more crucial in applications demanding real-time inference, such as robotic perception, UAV navigation, or safety-critical autonomous driving. Here, models must balance semantic expressivity with predictable latency and throughput, which pure LVLMs often fail to achieve without architectural compromises.

Recent hybrid approaches attempt to mitigate these trade-offs by decoupling semantic reasoning from spatial localization. Architectures like OV-DINO [172] and ContextDET [173] integrate large-scale pretrained LLMs for cross-modal reasoning, while retaining efficient transformer detectors (e.g., DINOv2 [174]) to handle box regression and classification. These systems aim to preserve open-vocabulary detection and compositional reasoning while lowering computational costs through modular design [175]. Similarly, models such as DetGPT employ a generate-then-detect framework, generating candidate objects or task-specific labels via LLMs and refining localization with lightweight detection heads [16]. However, such staged designs introduce additional complexity in integration and training, often requiring fine-tuned visual encoders or handcrafted alignment strategies [176,177]. Despite these innovations, current research lacks a systematic evaluation framework to quantify trade-offs in terms of FLOPs, latency, memory usage, and annotation cost per mAP gain. As LVLMs continue to evolve, future work should emphasize cost-effective scalability, model distillation, and architecture-aware benchmarking to bridge the gap between semantic capability and real-world usability.

[2] https://github.com/ultralytics/ultralytics.

### *4.5. Comparative insights into Open-Vocabulary Object Detectors (OVOD)*

OVODs such as ViLD [178], Grounding DINO [41], and OWL-ViT [179] represent a crucial middle ground between traditional object detectors and fully generative LVLMs. These models are uniquely designed to detect and localize arbitrary objects described by text prompts, enabling robust performance in unseen scenarios without retraining. ViLD pioneered the integration of CLIP-style embeddings into region proposal networks [180,181], effectively distilling vision-language representations into Mask R-CNN-style pipelines. Models like OWL-ViT and GLIP further advance this framework by combining joint vision-language pretraining with end-to-end detection fine-tuning [93]. Notably, YOLO-World [117] and YOLOE [118] preserve the computational efficiency of traditional YOLO architectures while integrating textual conditioning and prompt-driven detection capabilities.

OVODs consistently outperform traditional methods in zero-shot and domain-transfer settings. For example, Grounding DINO achieves an AP@50:95 of 48.3 on the RoadObstacle21 anomaly benchmark (Source Link), significantly outpacing standard YOLO in detecting out-of-distribution objects. Similarly, GLIP and YOLOE achieve high open-vocabulary recall while maintaining real-time inference speeds, making them suitable for safety-critical environments. Beyond accuracy metrics, OVODs offer enhanced adaptability through techniques such as dynamic prompt engineering exemplified by YOLO-World, which enables real-time vocabulary updates without retraining and modular architectures like LP-OVOD [182] and CCKT-Det, which support task-specific customization via lightweight modules, overcoming a core limitation of traditional fixed-vocabulary detectors. Moreover, models like OV-DETR [183] and OV-DINO [172] leverage transformer-based cross-attention or selective fusion to align visual and textual modalities more effectively than conventional detectors.

Despite their strengths, OVODs face several challenges. Models such as CCKT-Det [184] and Open Corpus OVD [185] show promising generalization when evaluated under corrupted input conditions such as fog, occlusion, or motion blur—but adversarial robustness remains inconsistent across architectures. While OWL-ViT demonstrates high tolerance to perturbations, most models still achieve limited performance in abstract reasoning tasks or in scenes requiring negation understanding. Another limitation is hallucination in prompt-conditioned scenes, where models may generate bounding boxes for implausible object relationships. Nevertheless, OVODs deliver superior zero-shot robustness, adaptability, and efficiency, making them valuable models for autonomous systems, remote sensing, and robotics where class boundaries are dynamic or undefined (see Table 5).

### *4.6. Evaluation metrics in LVLM-based object detection*

Evaluation of LVLMs in object detection tasks requires metrics that go beyond traditional precision–recall frameworks. Due to the inherent multimodality, open-vocabulary capacity, and compositional reasoning capabilities of LVLMs, specialized evaluation metrics and protocols have emerged.

Traditional metrics such as **mAP@0.5:0.95** remain essential for benchmarking spatial precision, but are insufficient to capture zero-shot performance or hallucination errors that may arise from vision-language misalignment [187]. Therefore, metrics like **Zero-Shot mAP**, **Hallucination Error Rate**, and **Open-Vocabulary Accuracy (OVA)** are critical for measuring generalization to novel classes and prompt understanding [97,188–190]. Additionally, metrics such as **CLIPScore** [191], **Compositional Error Rate** (e.g. in Taskclip [192]), and **Corruption mAP Drop** reflect semantic alignment, attribute binding failures, and robustness under distributional shifts [193].

Inference speed (**FPS**) and human alignment measures are also crucial for practical deployment in real-time or embodied environments. Moreover, tasks involving object counting or VQA require **MAE**, **RMSE**, and soft-accuracy based metrics. These metrics collectively form the quantitative framework for evaluating LVLMs in safety-critical, dynamic, and zero-shot object detection scenarios. Table 6 provides a comprehensive summary of key evaluation metrics tailored for LVLM-based object detection, detailing their mathematical formulations, definitions, and application contexts across zero-shot detection, semantic grounding, and robustness assessment.

## 5. Discussion

### *5.1. Discussion on current challenges and potential solutions*

Multimodal LVLMs mark a major advancement in object detection by integrating visual perception with natural language understanding. This integration enhances contextual reasoning, supports open-vocabulary recognition, and enables dynamic task interpretation crucial for applications such as robotics, autonomous navigation, and human–robot interaction. However, their real-world implementation remains limited because of several practical and architectural challenges. These challenges arise from the computational cost of processing large multimodal inputs, the difficulty of aligning linguistic prompts with spatial object regions, and the need for real-time inference in safety-critical systems such as autonomous vehicles or surgical robotics [70,211]. Additionally, ensuring robustness, reliability and generalization under noisy inputs, misaligned prompts, or domain shifts presents ongoing limitations [78].

To address these limitations, a systematic strategy is required. As depicted in Fig. 13a, resolving multimodal data complexity begins with region-aware pretraining and adversarial prompt tuning, progressing through architectural innovations like spatiotemporal encoders and decoupled prediction heads, ultimately enabling more effective context-aware detection. Complementing this model-level architectural strategy, Fig. 13b illustrates a ten-step roadmap encompassing efficiency optimization, prompt and fusion mechanisms, and reinforcement learning to enable real-time, scalable, and semantically rich object detection systems.

As shown by these strategies, it is important to emphasize the need for integrated solutions spanning model compression, data synthesis, hierarchical supervision, and modular fusion architectures. By addressing these areas, future LVLMs could become lightweight, robust, and interpretable systems capable of reliable detection in diverse environments. These figures collectively demonstrate how the field is progressing toward addressing current constraints while paving the way for future research in zero-shot learning, open-world object grounding, and multi-agent coordination.

The key current challenges in LVLM-based object detection are summarized as follows:

- **Computational Demands:** The deployment of LVLMs requires substantial computational resources, including large amounts of memory and access to high-performance GPUs, which can be prohibitive in resource-constrained environments [70,211].
- **Complex Integration:** Integrating diverse modalities such as visual data and natural language adds significant complexity to system design, challenging the synchronization of data streams and the alignment and fusion of features from different sources [212,213].

**Table 5**
Comparison of state-of-the-art Open-Vocabulary Object Detection (OVOD) models.

| Model | Architecture highlights | Speed (FPS) | mAP@50 | mAP @0.5:0.95 | Unseen class AP | Zero-shot capability | Unique advantages |
|---|---|---|---|---|---|---|---|
| ViLD [178] | CLIP distillation + Mask R-CNN with text prototypes | 7 | 72.2 | 36.6 | 29.1 | High (PASCAL VOC) | Zero-shot transfer w/o fine-tuning |
| OWL-ViT [179] | ViT backbone + joint image-text contrastive learning | 110 | 65.7 | – | – | Moderate | Adversarially robust, native prompt support |
| YOLO-World [117] | YOLO backbone + RepVL-PAN + text encoder | 161 | 68.7 | 21.2 | 38.5 | Moderate | Real-time prompt reparameterization |
| GLIP [186] | Unified detection + grounding via semantic alignment | 22 | 63.1 | 49.8 | 41.3 | Excellent | Phrase grounding + visual reasoning |
| YOLOE [118] | YOLO + RepRTA, SAVPE, LRPC modules | 130 | 67.3 | 52.6 | 44.2 | High | Prompt-free + multimodal features |
| LP-OVOD [182] | Linear probe using CLIP pseudo-labels | – | – | 40.5 | 34.9 | Moderate | Annotation-light + robust proposal filtering |
| OV-DETR [183] | DETR + conditional query-text alignment | 33 | – | 38.2 | 30.4 | Good | Cross-attention + modular queries |
| OV-DINO [172] | DINO + language-aware feature fusion | – | 47.3 | – | 42.6 | High | Real-time anomaly resilience |
| CCKT-Det [184] | Cyclic contrastive knowledge transfer + momentum encoders | – | – | – | 44.1 | High | Robust to corruption; long-tail generalization |
| Open Corpus OVD [185] | Web corpus prompts + region-based detection | – | – | 32.8 | 29.2 | Good | Adaptable to custom taxonomies |

**Table 6**
Metrics for evaluating LVLM-based object detection models.

| Metric | Definition/Formula | Use case/Notes |
|---|---|---|
| Zero-Shot mAP | $\text{mAP}_{\text{ZS}} = \frac{1}{N_{\text{unseen}}} \sum_{c \in \text{unseen}} \text{AP}_c$ | Measures generalization to novel classes without fine-tuning [97,188]. |
| mAP@0.5:0.95 | Mean average precision over IoU thresholds from 0.5 to 0.95 in steps of 0.05. | Penalizes loose bounding boxes; stricter than mAP@0.5 [194,195]. |
| Hallucination error rate | $\text{HER} = \frac{\text{FP}_{\text{relations}}}{\text{Total Predictions}} \times 100$ | Quantifies false positive relationships [196–198]; e.g., MERLIM benchmark reports 22% [199]. |
| Open-Vocabulary Accuracy (OVA) | *Human-rated correctness* on natural language queries. | GLIP [171] and GPT-4V (Source Link) outperform traditional detectors on complex prompts. |
| Frames Per Second (FPS) | $\text{FPS} = \frac{\text{Frames Processed}}{\text{Total Time (sec)}}$ | Real-time capability measure. E.g., YOLO-World achieves 161 FPS [117]. |
| CLIPScore | $\text{CLIPScore} = \cos\left(\text{CLIP}_{\text{img}}(I), \text{CLIP}_{\text{text}}(T)\right)$ | Evaluates alignment of generated text with image; >0.8 indicates strong grounding [200–202]. |
| Counting MAE/RMSE | $\text{MAE} = \frac{1}{N}\sum_{i=1}^{N} \lvert y_i - \hat{y}_i \rvert \quad \text{RMSE} = \sqrt{\frac{1}{N}\sum_{i=1}^{N}(y_i - \hat{y}_i)^2}$ | Used in LVLM-Count; Lower MAE/RMSE indicates better counting accuracy [203–205]. |
| VQA accuracy | *Soft match accuracy* (accepts synonyms or rephrasings). | Benchmark: MM-Ego [206] with 67.3% on egocentric QA. [206,207] |
| Compositional error rate | Failure in object-attribute bindings in compositional prompts. | Measured using synthetic scenes (e.g., “red cube on blue sphere”) [208–210]. |

- **Training Data Requirements:** Effective training of LVLMs demands extensive and diverse datasets, which are costly and labor-intensive to compile [80]. These datasets must include accurate annotations across multiple modalities (e.g., visual, textual), adding another layer of complexity to their preparation.
- **Inference Speed:** The complex architectures and large size of LVLMs contribute to slower inference speeds [71,92,159,160], making them less suitable for applications requiring real-time decision-making, such as autonomous driving or interactive robotics.
- **Robustness and Generalization:** While LVLMs excel at handling tasks with open vocabulary and can interpret contextual cues, they are susceptible to issues like prompt dependency and may produce hallucinated outputs [156,211]. This can undermine their effectiveness in scenarios where accuracy and reliability are critical.
- **Domain Gap Between Pre-training and Detection Tasks:** There is a clear mismatch between the image-level supervision used during the pre-training of VLMs and the region-level precision required for object detection tasks [80,211]. This gap can significantly impact the performance of LVLMs when applied to specific detection scenarios.
- **Image-Level vs. Region-Level Understanding:** VLMs like CLIP, designed for global image understanding [131,187], face performance deterioration when tasked with the localized analysis necessary for object detection, resulting in a loss of contextual accuracy [156,159,160].
- **Background Class Representation Challenge:** Unlike traditional object detection models, LVLMs lack a dedicated representation for “background”, leading to misclassifications and increased false positives [187,211].
- **Contextual Information Loss:** The application of LVLMs to localized regions can result in the loss of essential contextual information [11,211], which is critical for the accurate classification of objects within their environment.

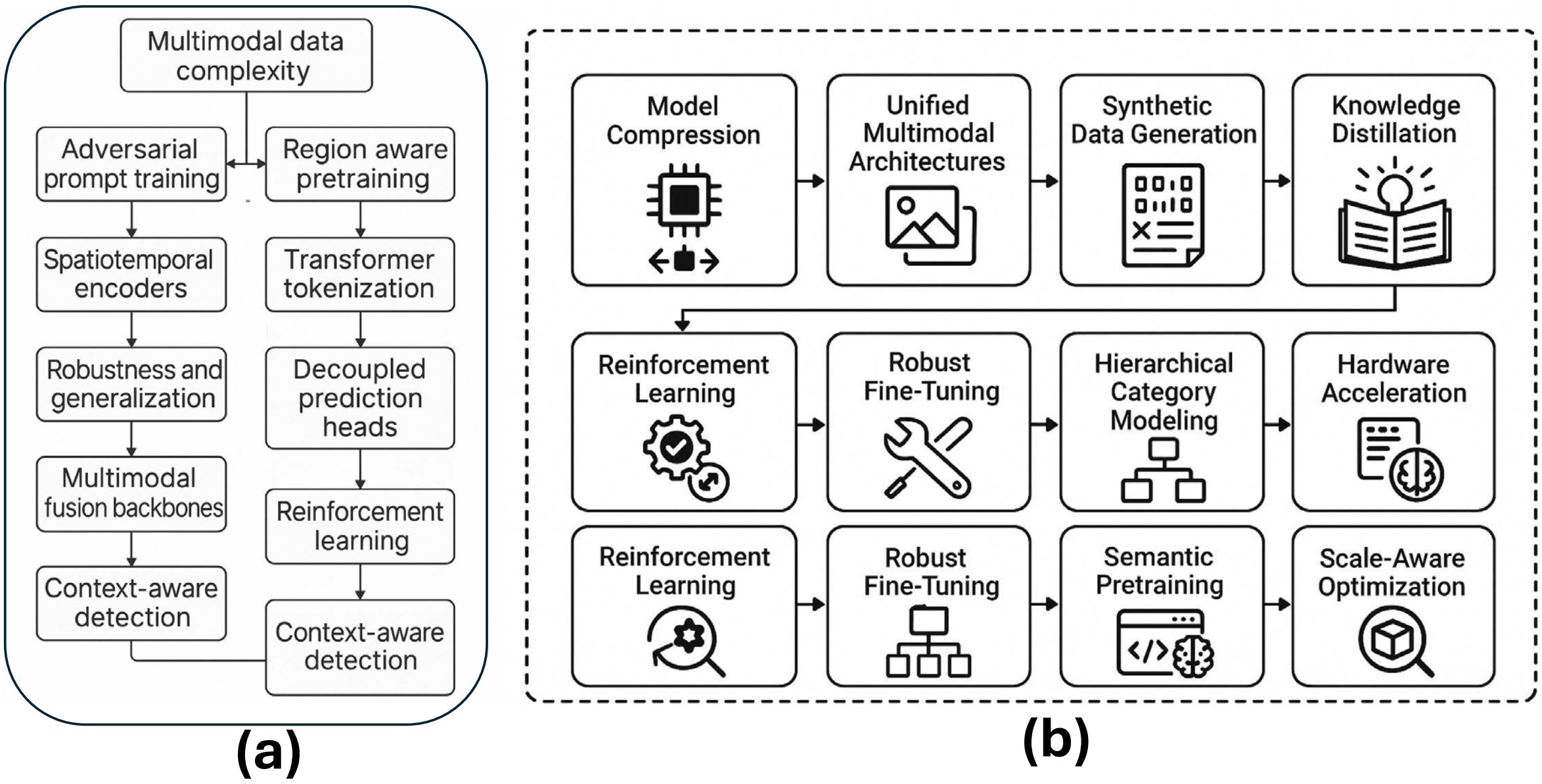

**Fig. 13.** Strategic roadmap for object detection with LVLMs. (a) This flowchart proposes a pipeline to overcome key limitations in LVLM-based object detection. It begins by addressing multimodal data complexity through region-aware pretraining and adversarial prompt tuning. These enhanced cross-modal representations feed into spatiotemporal encoding and transformer-based tokenization modules, leading to modules focused on robustness, decoupled predictions, and reinforcement learning. The system converges into multimodal fusion backbones, enabling context-aware and reliable object detection in dynamic environments; (b) This info-graphic illustrates ten major future pathways aimed at enhancing LVLM performance. It highlights architectural innovations, scalable training strategies, advanced prompt engineering, temporal modeling, and open-vocabulary adaptation. The figure presents a unified roadmap for lightweight, generalizable LVLMs, enhancing multimodal fusion, reasoning, and real-time object detection capabilities.

- **Noisy Pseudo-Label Generation and Error Accumulation:** The reliance on generating pseudo-labels for novel categories introduces errors, particularly noisy boxes and biases [187,211], which are further amplified during the training of detection models.
- **Mislocalization Issues:** The imprecise localization of objects by LVLMs [131], coupled with equal loss weighting during training, results in degraded detection quality, particularly for novel categories [187].
- **Base-Novel Category Conflicts:** The simultaneous training to recognize both seen and unseen categories leads to label assignment conflicts and necessitates a delicate balance to optimize detection across base and novel categories without sacrificing accuracy [120,136].
- **Semantic Boundary Challenges:** LVLMs must navigate the fuzzy semantic boundaries between overlapping or hierarchically related categories [136,187], which complicates the distinction between base and novel categories in object detection tasks.

While recent LVLM-based object detection methods acknowledge key limitations such as hallucination, semantic misalignment, and spatial mislocalization, few offer solutions or mechanisms for their mitigation. Example studies partially addressing hallucination, where models falsely detect objects due to ambiguous language cues or multimodal mismatch, are Grounding DINO [42] and GLIP [171]. These studies propose contrastive alignment and grounding losses as the solutions for hallucination, but are typically evaluated under curated benchmarks and lack robust testing across dynamic, compositional prompts, as evidenced by MERLIM's reported 22% error rate [199]. Similarly, models such as ContextDET [173] attempt to improve semantic grounding via generate-then-detect pipelines, but the method struggles with negation and abstract reasoning due to limitations in their language modeling depth and training data diversity. Mislocalization issues are tackled by hybrid methods (e.g., DetGPT [16], OV-DETR [183]) that combine coarse VLM priors with conventional detectors, but these methods often rely on handcrafted thresholds or additional modules that are hard to generalize across domains [214,215]. Consequently, despite these innovations, many proposed solutions remain narrow in scope and lack systematic evaluations under real-world perturbations, such as occlusion, illumination shifts, or adversarial prompts. This underscores the need for benchmark expansions and robustness-centric design strategies in future LVLM research.

### *5.2. Strategic directions and research outlook for LVLM-based object detection*

The key challenges, potential future solutions, and their anticipated impact on improving LVLM-based object detection are further summarized in Table 7 and in the following five-points providing a structured analysis across these critical dimensions.

- **Towards Efficient and Scalable LVLM Deployment:** In the future, reducing the memory and GPU demands of LVLMs could be achieved through techniques such as quantization [326], pruning [327], model distillation [142], and LoRA-based tuning [196,328]. These model compression and adaptation techniques would allow LVLMs to operate efficiently on edge devices and in resource-constrained environments. LoRA (Low-Rank Adaptation), in particular, enables parameter-efficient fine-tuning by updating only small trainable matrices within transformer layers. Additionally, introducing cascaded models and early-exit mechanisms could dynamically adjust inference depth, enabling real-time object detection with significantly lower latency and computational resources.
- **Advancing Multimodal Fusion and Localized Reasoning:** To overcome the integration complexity of visual and textual modalities, future architectures could employ unified multimodal encoders with spatiotemporal attention [329], such as the Perceiver or hybrid fusion backbones [330,331]. For precise region-level understanding, transformer-based regional tokenization, hierarchical reasoning layers, and the inclusion of background-aware representations could significantly enhance detection accuracy [332]. Context-aware heads and scene graph integration could ensure that even localized object detection preserves holistic scene semantics [333,334].

**Table 7**
Summary of future directions for overcoming key challenges in LVLM-based object detection. It presents critical limitations, outlines targeted solutions, including architectural, data-centric, and training innovations, and details expected impacts across application domains, such as robotics, medical imaging, remote sensing, and real-time analytics.

| Challenge area | Potential solution | Expected impact |
|---|---|---|
| Computational resource constraints | Model quantization [216], transformer pruning [217,218], knowledge distillation [219,220], and efficient LVLM architecture design [221] | Enables LVLM deployment on edge devices and mobile platforms [213,222,223] |
| Multimodal integration complexity | Design of unified cross-modal fusion modules [131,224] and temporal-spatial synchronization mechanisms [225] | Improves multimodal reasoning accuracy in indoor robot navigation [148,226]; Reduces alignment errors in autonomous vehicles [141]; Enables coherent processing of streaming vision-text data in augmented reality systems [129] |
| Data scarcity and annotation cost | Synthetic data generation via GPT-4 + SAM [67,227,228], few-shot transfer learning [103], multimodal data augmentation [229,230] | Decreases dependence on costly manual annotations in object detection [67,227]; Expands training scalability for underrepresented object categories [129]; Facilitates cross-domain transfer in imaging with visual-textual descriptions such as medical imaging [231] |
| Slow inference speed | Hardware-aware pruning [232,233], lightweight attention variants [63], and real-time transformers [234] | [235,236]; Improves latency in real-time autonomous decision-making [237] |
| Robustness and prompt sensitivity | Prompt tuning [238], multi-context reasoning modules [239], and uncertainty-aware decoding strategies [240,241] | Minimizes hallucinations in zero-shot detection across various imageries [196,242]; Improves fault tolerance in different systems such as medical diagnostics [243] |
| Pretraining vs. detection domain gap | Introduce region-level pretraining objectives [244–246] and detection-specific visual-language heads [120] | Aligns pretraining with object detection outputs in various images (e.g., satellite imagery) [129]; Improves bounding box precision for automation operatopms [129,141]; Minimizes performance degradation when transferring models across different object detection tasks. [160] |
| Image vs. region understanding | Implement localized attention with global scene context fusion [11] | Strengthens object localization for applications such as traffic monitoring [141,221]; Enables detailed object extraction in visual search [93]; Balances semantic abstraction with pixel-level fidelity [247] |
| Background representation challenge | Introduce explicit background class embeddings [64,248] and adaptive contrastive learning [131] | Improves foreground-background separation in surveillance [249]; Decreases false positives in monitoring [250]; Enhances scene parsing in mixed-reality navigation [251] |
| Contextual information loss | Spatial memory networks [252–254], global-local context encoders [255,256], and scene graph encoders [257] | Preserves object-scene interactions, such as those in home robotics [258]; Boosts classification accuracy in crowded scenes [160]; Enhances object reference resolution in vision-language tasks, e.g., VQA and visual grounding [259]. |
| Noisy pseudo-labels and error accumulation | Label cleaning via self-training [260,261], spatiotemporal consistency checks [262], and ensemble consensus filtering [263,264] | Mitigates overfitting to noisy supervision in complex domains (e.g., wildlife monitoring) [265]; Stabilizes label refinement in iterative self-training setups (e.g., pseudo-label bootstrapping) [266,267]; Enhances recognition of rare classes in imbalanced datasets (e.g., long-tail object categories) [268]; |
| Mislocalization and loss imbalance | Use of dynamic loss scaling [221,269], spatial attention refinement heads [270,271], and IoU-guided supervision [272,273] | Enhances object boundary accuracy in aerial mapping [269]; Improves small-object detection [152]; Balances focus on rare vs. frequent object types [100] |
| Base-novel category conflict | Task-balanced training [274] and embedding disentanglement with class-conditional prompts [275] | Boosts zero-shot generalization in industrial inspection [276]; Reduces bias toward frequent training classes [276]; Improves calibration across base and novel categories [152] |
| Semantic boundary challenges | Incorporate ontology-guided supervision [277,278], soft taxonomy-aware classifiers [279], and class hierarchy constraints [160] | Improves fine-grained object recognition [266]; Reduces confusion in hierarchical categories like animals vs. pets [280]; Supports structured prediction in scientific image analysis [129] |
| Task-specific prompt engineering limitations | Automated prompt generation using instruction-tuned LLMs [281], reinforcement-based refinement [282], and [257] | Enhances model adaptability in diverse object detection contexts (e.g., warehouse robotics) [90]; Reduces human effort and variability in prompt design [97]; Improves consistency and interpretability across multimodal queries [11] |
| Lack of temporal awareness in dynamic scenes | Temporal fusion modules [221,249], video-VLM pretraining [283], and sequential attention for motion-aware object grounding [284–286] | Enables tracking-aware detection in video surveillance and autonomous driving [179]; Captures object transitions for better scene understanding [90]; Enhances performance in spatiotemporal tasks like action-object detection [287] |
| Scalability to open-world object categories | Incremental learning [160,288,289], open-vocabulary expansion using weak supervision [290,291], and knowledge graph grounding [292,293] | Expands LVLM coverage to rare or newly introduced object classes [160]; Reduces retraining needs when new categories are added [103,129] |

**Table 8**
Emerging evaluation benchmarks for Vision-Language Models (VLMs). This table highlights diverse tasks and evaluation criteria, covering object detection, hallucination, reasoning, fine-grained understanding, and emotional comprehension.

| Benchmark | Primary focus | Key metrics | Evaluation criteria | Highlights/Notes |
|---|---|---|---|---|
| Roboflow100-VL [294] | Multi-domain object detection (100 datasets; 564 classes, 164K images) | Zero-shot mAP | Zero-shot/few-shot mAP, domain adaptation, class diversity | Exposes poor zero-shot generalization for VLMs; covers medical, satellite, industrial domains. Best zero-shot: GroundingDINO (15.7 mAP) [295]. |
| MiniGPT-4 [296] | Caption-guided object localization and grounding (COCO) | mAP@0.5, mAP@0.50:95 | Instruction alignment, grounding precision, clutter resistance | Excels in instruction following; struggles with occlusion and scene clutter [170,297,298]. |
| mPLUG-Owl [299] | Open-vocabulary detection and cross-modal understanding (COCO, LVIS) | mAP@0.5, mAP@0.50:95 | Zero-shot classification, VQA accuracy, grounding fidelity | Strong on long-tail categories; reduced accuracy in dense layouts [300,301]. |
| MM-ReAct [302] | Real-time multimodal reasoning and action (COCO, ScienceQA) | mAP@0.5, mAP@0.50:95 | Inference latency, interaction throughput, reasoning complexity | Dialogue-based, real-time capable [303]; leverages LLMs for planning and vision experts for perception [304]. |
| GPT-4V + SAM [305] | Referring expression segmentation (RefCOCO, RefCOCOg) | mAP@0.5, mAP@0.50:95 | Referential comprehension, multimodal consistency, segmentation quality | Robust in interactive referring tasks; combines GPT-4V's comprehension with SAM's segmentation [170]. |
| Vision-LLM [120] | Egocentric vision and task-oriented detection (Ego4D) | mAP@0.5, mAP@0.50:95 | First-person task accuracy, AR suitability, real-time adaptability | Strong for egocentric robotics/AR; limited open-vocabulary performance [306]. |
| MM-Ego [206] | Egocentric video QA and memory (Ego4D, 7M QA pairs) | Video QA accuracy | Long-horizon memory, detail retention, bias mitigation | 629 videos, 7026 questions; introduces memory pointer prompting for extended content understanding [307,308]. |
| MERLIM [309] | Object recognition, counting, and compositional bias | Hallucination error rate | Hallucination error rate, relational accuracy, compositional bias | Exposes hallucinations and compositional errors in object relationships [199,310]. |
| Video OCR benchmark [311] | Scene-text recognition in dynamic video streams (1477 frames) | WER, CER | Word Error Rate (WER), Character Error Rate (CER), occlusion robustness, spatiotemporal coherence | VLMs (GPT-4o [312], Gemini-1.5 [313], Claude-3 (Source Link) outperform classic OCR in dynamic settings; challenges remain for stylized/occluded text [314]. |
| Open-ended VQA benchmark [315] | Visual reasoning via follow-up semantic queries (classification datasets) | Reasoning coherence | Taxonomy-guided reasoning, chain-of-thought coherence, label consistency | LLM-based VLMs align closely with human logic in layered VQA; uses semantic label hierarchies [316–318] |
| Real-world error understanding [319] | Logical, temporal, and factual error detection across scenes | Error score | Qualitative error scoring, human alignment ranking | GPT-4V identifies dynamic scene inconsistencies; surpasses LLaVA [320] and Qwen-VL [321] in human ratings. |
| FG-BMK [322] | Fine-grained object understanding and feature sensitivity (3.49M Qs, 3.32M images) | Accuracy, mAP (retrieval) | Semantic accuracy, attribute sensitivity, perturbation robustness | Human- and machine-oriented paradigms; reveals model blind spots to fine-grained features and perturbations [323]. |
| EasyARC [324] | Multi-step visual reasoning and pattern induction (procedural ARC tasks) | Success rate | Task success rate, self-correction, abstraction depth | Procedurally generated, scalable; tests multi-image, multi-step reasoning and RL suitability [324]. |
| EmoNet-Face [325] | Fine-grained facial emotion recognition (40 emotions, 2500 expert-annotated images) | Accuracy, F1 | Emotion classification accuracy, robustness, human error explainability | 40-category taxonomy, expert annotations, demographic balance; sets new standard for affective VLM evaluation [325]. |

- **Reinforcement Learning and Reward Design for Fine-Grained Supervision:** Reinforcement learning could play a vital role in fine-tuning LVLMs for better localization and label precision [335]. In the future, customized reward functions such as odLength could mitigate reward hacking by penalizing excessive predictions, leading to more robust and reliable detections. Curriculum-based RL [336], confidence-weighted loss functions [151,242], and iterative relabeling [337] could stabilize learning from noisy pseudo-labels, especially in open-vocabulary and zero-shot detection tasks [338].
- **Improving Robustness, Generalization, and Semantic Disambiguation:** Future LVLMs could reduce hallucinated outputs and prompt dependency through adversarial prompt training [196, 339], uncertainty modeling [340], and retrieval-augmented reasoning pipelines [341]. The domain gap between image-level pretraining and region-level detection could be addressed by designing pretraining objectives that include region-aware contrastive losses and adapter modules fine-tuned on detection tasks. Base-novel conflicts could be mitigated using decoupled prediction heads and label-space-aware balancing mechanisms, while hierarchical category modeling could help disambiguate fuzzy semantic boundaries [342,343].
- **Leveraging Data Complexity and Model Scaling Strategically:** To better exploit the reasoning capacity of LVLMs, future systems could train on semantically complex and richly annotated datasets like D3 rather than simple category labels in COCO [260,344]. This shift would encourage stronger reasoning chains during detection. Furthermore, reinforcement learning could be tailored to different model sizes, as larger models like 7B and 32B exhibit more pronounced gains in reasoning-intensive tasks [238, 269]. Such scale-aware optimization strategies could maximize the benefit of RL in generalization across both seen and unseen categories [275,334].

### 5.3. Quantitative benchmark comparison of LVLMs

Benchmark datasets such as COCO (Source Link), LVIS (Source Link), and custom domain-specific datasets (e.g., RefCOCO (Source Link)), Ego4D (Source Link) play a critical role in evaluating the capabilities of multimodal LVLMs in object detection. These benchmarks differ in their object category granularity, scene complexity, and annotation richness, offering complementary insights into model performance. For example, COCO emphasizes diverse everyday scenes and object localization; LVIS focuses on fine-grained categories and long-tail distributions; RefCOCO targets referring expression comprehension, and Ego4D involves egocentric, action-based object recognition. Table 8 provides a comparative overview of five state-of-the-art LVLMs evaluated in terms of detection accuracy (mAP@0.5 and mAP@0.5:0.95), inference speed (FPS), and notes on generalization ability. These results highlight the trade-offs between performance and deployment feasibility across real-time, robotics, and cross-modal reasoning applications.

### 5.4. Impact of LVLM-based object detection on future of robotics

The future of object detection using Multimodal Large Vision-Language Models (LVLMs) lies in their ability to effectively fuse visual perception with semantic understanding across open-world settings. These models enable flexible and scalable detection by interpreting both visual inputs and language instructions, making them applicable across various domains. In robotics, LVLMs are increasingly used to facilitate visual reasoning in dynamic tasks such as home assistance, warehouse automation, and human–robot interaction. For instance, vision-language integration empowers service robots to identify and fetch objects based on verbal commands, or allows industrial robots to adapt to changing environments without retraining. While the broader value of LVLMs is in their potential to advance general-purpose, open-vocabulary object detection in real-world, multimodal environments, their capabilities enable robotics as an compelling and impactful application area.

GR00T N1 [345] and Helix[3] as illustrated in the Fig. 14 is a cutting-edge example of how LVLMs can be integrated into robotic systems to enhance their perceptual and cognitive capabilities. Its VLA model combines a vision-language module that processes visual and textual inputs to understand and interpret the environment, with a Diffusion Transformer module that generates precise motor actions in real time. This integration allows the robot to perform tasks that require both high-level cognitive functions and fine-motor execution, such as navigating complex environments and manipulating objects in ways that were previously challenging for automated systems.

The adaptability and open-vocabulary capabilities of LVLMs, as detailed in Table 4, allow robots like GR00T N1 to operate effectively in varied and unforeseen scenarios without needing retraining for every new object or task. This capability is crucial for real-world applications such as household, healthcare and agriculture where unpredictability is common. The robot's ability to interpret and act upon language instructions in real-time, leveraging the multimodal data, aligns closely with the needs of next-generation robotic systems designed for generalist roles in human environments.

Moreover, LVLMs enable robots to understand context better, make informed decisions, and learn from minimal data, echoing the capabilities necessary for generalist humanoid robots. As depicted in Fig. 14, GR00T N1's model architecture highlights the seamless integration of vision and language understanding with dynamic action generation, setting a standard for future developments in robot design.

In the future, it is expected that there will be greater integration of LVLMs into various aspects of robot functionality. Future developments may focus on enhancing the efficiency and speed of LVLMs to meet the demands of real-time processing and task execution, reducing the computational overhead, and expanding the models' capabilities to handle more complex, multi-step tasks autonomously. Additionally, as robots become more embedded in daily tasks, the ability of LVLMs to process and understand multimodal human-centric data will be crucial for developing robots that can adapt to and learn from their interactions with humans and their environments.

In essence, the evolution of multimodal LVLMs and their integration into robotics exemplified by systems such as GR00T N1 and Helix by Figure marks a transformative step toward the development of more autonomous, context-aware, and intelligent robotic agents. These advanced models do not merely enhance perception or language comprehension in isolation; they enable a deeper fusion of multimodal reasoning, allowing robots to interpret nuanced, uncertain environments, follow complex instructions, and make informed decisions in real time. This synergy between vision and language is increasingly critical for deploying robots in real-world scenarios that demand flexible cognition and adaptive behavior. These advancements are not only expanding the operational capacities of robots but also paving the way for their adoption in domains previously considered too ambiguous, unstructured, or dynamic for automation. For instance, in 'elderly care', robots must interpret both visual cues and spoken language to assist with medication reminders, object retrieval, or social interaction tasks that require a rich understanding of both context and intention. In 'assisted cooking', robots must recognize ingredients, interpret natural-language recipes, and adapt to varied kitchen layouts. 'Disaster response' is another important application, where robots navigate unstable and uncertain environments, interpret commands in noisy conditions, and visually identify victims or hazards. 'Interactive teaching and tutoring' for children and neurodivergent individuals also benefit from multimodal understanding, requiring the ability to detect engagement, interpret questions, and provide contextualized, visual explanations. In all these domains, among others, LVLMs serve as the cognitive backbone, enabling robotic systems to bridge the gap between perception and action an essential capability for the next generation of real-world, general-purpose robots.

## 6. Conclusion

Object detection has long been a cornerstone task in computer vision, with traditional machine learning methods like SVMs and hand-crafted features giving way to deep learning architectures such as YOLO, Mask R-CNN, Faster R-CNN, and detection transformers (DETRs). These models have achieved remarkable performance in real-time localization and classification tasks across various domains. However, the recent emergence of LVLMs introduces a transformative paradigm by integrating natural language understanding with visual perception, enabling more context-aware, generalizable, and semantically rich object detection capabilities.

In this first known review on this topic, we evaluated the state-of-the-art developments and provided an in-depth examination of the architectural innovations in Multimodal LVLMs for object detection. This study not only highlights the key architectural improvements and methodologies of LVLMs but also presents a comprehensive comparison against conventional models such as YOLO and Faster R-CNN. It was found that while LVLMs excel in contextual understanding and multimodal interactions, traditional frameworks remain important for applications requiring high-speed, precision, and real-time processing on edge devices. Our analysis, therefore, demonstrates the complementary nature of LVLMs to traditional object detection systems. The integration of NLP and computer vision in LVLMs opens up new avenues for enhancing scene understanding and automating preliminary tasks such as labeling, while dedicated CV models continue to manage precise, real-time localization tasks. We discussed the foundational functioning and evolution of multimodal LVLMs, revealing how these technologies merge to advance vision tasks through intuitive language-driven interfaces and detailed the inherent challenges and limitations of these systems. Looking ahead, LVLMs and traditional deep learning models will likely coexist, each reinforcing the other's core strengths. This integration promises to expand the capabilities of object detection systems, making them more adaptive, efficient, and accessible across various scenarios. In our analysis, we identified major challenges such as high computational costs, prompt dependency, noisy pseudo-labels, noisy pseudo-labels, and the domain gap between pretraining and detection. To address these, we proposed solution including region-aware pretraining, model compression, reinforcement learning, and improved multimodal fusion. e believe this review serves as a foundational document, critically analyzing the current landscape of LVLM-based object detection and setting the stage for future innovations in this rapidly evolving field of automation and robotics, where contextual reasoning and spatial precision must be harmoniously integrated.

[3] https://www.figure.ai/.

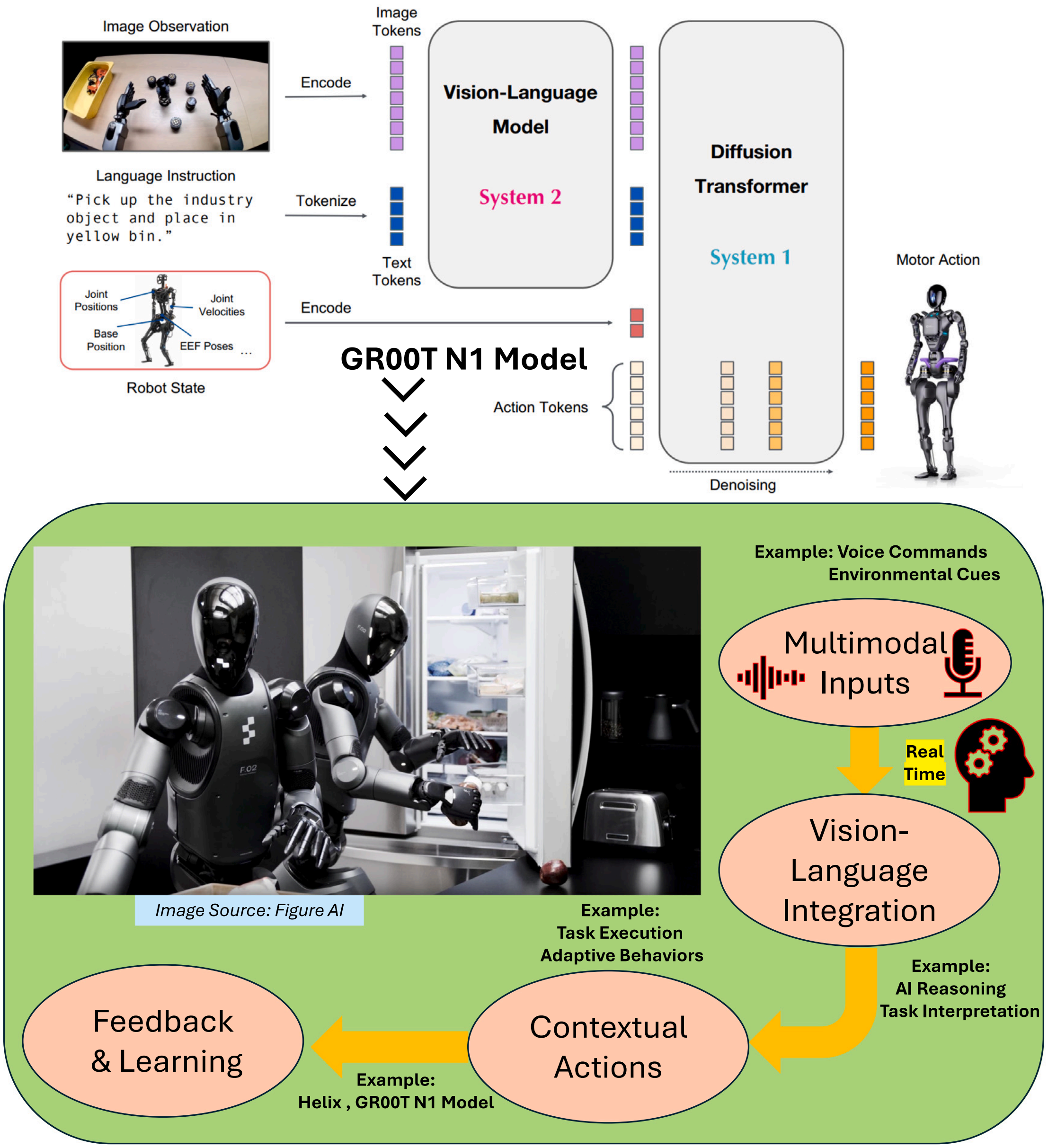

**Fig. 14.** Illustrating the future prospect of object detection for robotic applications, leveraging advancements in multimodal LVLMs. The upper portion of the diagram illustrates the NVIDIA GR00T N1 model [345], a cutting-edge Vision-Language-Action (VLA) system. In this model, multimodal inputs image observations and language instructions are transformed into tokens processed by the LVLM's backbone. These tokens, integrated with robot state and action encodings, facilitate the generation of precise motor actions via a Diffusion Transformer module. The lower section of the figure delineates the envisioned progression through four pivotal stages: Multimodal Inputs, Vision-Language Integration, Contextual Actions, and Feedback & Learning, highlighting the seamless integration of AI in enhancing robotic capabilities and responsiveness in dynamic environments.

## CRediT authorship contribution statement

**Ranjan Sapkota:** Writing – review & editing, Writing – original draft, Visualization, Methodology, Investigation, Formal analysis, Conceptualization. **Manoj Karkee:** Writing – review & editing, Writing – original draft, Visualization, Resources, Project administration, Methodology, Investigation, Funding acquisition, Formal analysis.

## Statement on AI Writing Assistance

ChatGPT and Perplexity were utilized to enhance grammatical accuracy and refine sentence structure; all AI-generated revisions were thoroughly reviewed and edited for relevance. Additionally, ChatGPT-4o was employed to generate realistic visualizations.

## Declaration of competing interest

The authors declare that they have no known competing financial interests or personal relationships that could have appeared to influence the work reported in this paper.

## Acknowledgment

This work was supported in part by the National Science Foundation (NSF) and the United States Department of Agriculture (USDA), National Institute of Food and Agriculture (NIFA), through the "Artificial Intelligence (AI) Institute for Agriculture" program under Award Numbers AWD003473 and AWD004595, and USDA-NIFA Accession Number 1029004 for the project titled "Robotic Blossom Thinning with Soft Manipulators." Additional support was provided through USDANIFA Grant Number 2024-67022-41788, Accession Number 1031712, under the project "ExPanding UCF AI Research To Novel Agricultural EngineeRing Applications (PARTNER).".

## Data availability

Data will be made available on request.